\PassOptionsToPackage{dvipsnames}{xcolor}
\documentclass{fairmeta}

\newif\ifarxiv\arxivtrue

\usepackage{amsmath}
\usepackage{amsfonts}
\usepackage{enumitem}
\usepackage{xspace}
\usepackage{siunitx}
\usepackage{nicefrac}
\usepackage{url}
\usepackage{listings}
\usepackage{palatino}
\defcitealias{faircodegenteam2025cwmopenweightsllmresearch}{CWM}
\defcitealias{cohere2025commanda}{Command A}

\newcommand{\matthijs}[1]{{\color{OliveGreen} [\textbf{MD}: #1]}}
\newcommand{\pem}[1]{{\color{cyan} [\textbf{PE}: #1]}}
\newcommand{\gsz}[1]{{\color{Aquamarine} [\textbf{G}: #1]}}
\newcommand{\maria}[1]{{\color{red} [\textbf{ML}: #1]}}
\newcommand{\daniel}[1]{{\color{cyan} [\textbf{D}: #1]}}
\newcommand{\rv}[1]{{\color{blue} [\textbf{H}: #1]}}
\newcommand{\manu}[1]{{\color{violet} [\textbf{MF}: #1]}}
\newcommand{\scott}[1]{{\color{teal} [\textbf{SY}: #1]}}

\iftrue
\renewcommand{\matthijs}[1]{{}}
\renewcommand{\pem}[1]{{}}
\renewcommand{\gsz}[1]{{}}
\renewcommand{\maria}[1]{{}}
\renewcommand{\daniel}[1]{{}}
\renewcommand{\rv}[1]{{}}
\renewcommand{\manu}[1]{{}}
\renewcommand{\scott}[1]{{}}
\fi

\title{Self-Pruned Key-Value Attention: Learning \newline When to Write by Predicting Future Utility}

\author[*,1]{Gergely Szilvasy}
\author[*,1,2]{Manuel Faysse}
\author[1]{Maria Lomeli}
\author[1]{Matthijs Douze}
\author[1]{\newline Pierre-Emmanuel Mazaré}
\author[1]{Loïc Cabannes}
\author[1]{Wen-tau Yih}
\author[1]{Hervé Jégou}

\affiliation[1]{Meta FAIR}
\affiliation[2]{MICS, CentraleSupélec}

\contribution[*]{Equal contribution}
\correspondence{\email{gsz@meta.com}}
\date{\today}

\abstract{

Under modern test-time compute and agentic paradigms, language models process ever-longer sequences. Efficient text generation with transformer architectures is increasingly constrained by the Key-Value cache memory footprint and bandwidth. 
To address this limitation, we introduce \emph{Self-Pruned Key-Value Attention} (SP-KV), 
a mechanism designed to predict future KV utility in order to reduce the size of the long-term KV cache. This strategy operates at a fine granularity:  a lightweight utility predictor scores each key-value pair, and while recent KVs are always available via a local window, older pairs are written in the cache and used in global attention only if their predicted utility surpasses a given threshold.
The LLM and the utility predictor are trained jointly end-to-end exclusively through next-token prediction loss, and are adapted from pretrained LLM checkpoints.

Rather than enforcing a fixed compression ratio, SP-KV performs \emph{dynamic} sparsification: the mechanism adapts to the input and typically  reduces the KV cache size by a factor of $3$ to $10\times$, longer sequences often being more compressible. This leads to vast improvements in memory usage and decoding speed, with little to no degradation of validation loss nor performance on a broad set of downstream tasks. 
Beyond serving as an effective KV-cache reduction mechanism, our method reveals structured layer- and head-specific sparsity patterns that we can use to guide the design of hybrid local-global attention architectures.

}

\begin{document}

\maketitle

\begin{figure}[h!]
    \centering
    \begin{minipage}{0.61\linewidth}
    \raisebox{7pt}{
    \includegraphics[width=\linewidth]{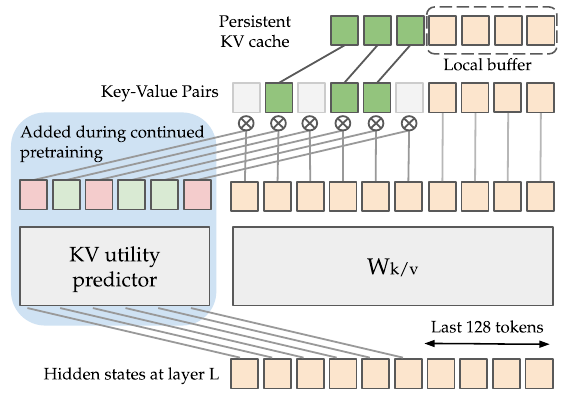}}
    \end{minipage}
    \hfill
    \begin{minipage}{0.36\linewidth}
    \captionof{figure}{
    \textbf{Overview of Self-Pruned Key Value Attention}: The learned KV utility predictor conditions key-value utilization in the attention operation.
    At inference, only KV pairs above a given utility threshold $\tau$ are kept in the persistent KV cache, enabling memory savings and decoding speedups.
    We always keep the recent past (128 tokens) to preserve local interactions.
    During training, token selection is replaced by differentiable gating to preserve gradient flow.
    Models pretrained with full attention progressively sparsify under continued pretraining without the need of a specific loss.
    \label{fig:overview}}
    \end{minipage}
\end{figure}

\section{Introduction}

At inference, the size and speed of transformer language models~\citep{Vaswani2017AttentionIA, brown2020language} are increasingly limited by memory rather than compute. 
During autoregressive generation, their key--value (KV) cache grows linearly with sequence length and is read by every newly generated token. 
As deployments shift toward long-context, retrieval-augmented, and agentic test-time pipelines, this expanding cache turns GPU memory traffic into a central performance bottleneck. 
This pressure extends to post-training, where long-context reinforcement learning and tool-integrated agent training rely on extended decoding rollouts~\citep{zhu2025scalingttcagents,wang2025loongrl}.

To mitigate the issue, architectural approaches such as GQA ~\citep{ainslie2023gqa} or MLA  ~\citep{liu2024deepseekv2} reduce the KV cache size by sharing keys across query head groups. 
Other approaches exploit the fact that most query-key interactions are concentrated within a short local window, whereas long-range interactions are sparser \citep{zhang2023h2o}. 
Typically, hybrid transformers reduce their reliance on global attention by interleaving the usual global attention with local sliding-window attention~\citep{beltagy2020longformer,riviere2024gemma2}, or by replacing certain attention layers with fixed-memory sequence mechanisms such as Gated DeltaNet~\citep{yang2025gateddeltanet}.
A separate line of work exploits \emph{read-time sparsity}: query-aware methods such as QUEST and DeepSeek Sparse Attention retrieve only a subset of past keys during decoding. While speed is improved, the full cache is kept in memory~\citep{tang2024questqueryawaresparsityefficient,deepseekai2025deepseekv32pushingfrontieropen}. 
In this paper, we rely on the observation that, for a given attention head, queries and keys mostly specialize into short and long-term interactions, suggesting that only a subset of past key--value pairs is consistently useful for future decoding.  
This raises a question: 
\begin{center}
\textit{Is it necessary to write every token indiscriminately into long-term/persistent memory?}
\end{center}

If no, it implies that the KV Cache could be \emph{sparsely read}, but crucially also that we could \emph{selectively write} into it, thereby saving memory in addition to FLOPs.
Previously, eviction methods such as H$_2$O and KVZap have attempted to prune the cache after prefill using past token statistics or learned policies applied to a frozen model~\citep{zhang2023h2o,jegou2026kvzap}. 
Although these techniques can substantially reduce memory usage, they often do so at the expense of model quality. A central limitation of post-hoc methods is that the model's internal token representations are not adapted to the pruning strategy, while the pruning mechanism itself is typically calibrated on small auxiliary datasets. This leads to a train--test mismatch that worsens as compression becomes more aggressive and as the input distribution shifts (see Section~\ref{sec:rw}). 
We make the following contributions: 

\fcolorbox{gray!10}{gray!10}{
\begin{minipage}{0.975\linewidth}\noindent\textbf{Contribution 1.} We introduce Self-Pruned KV Attention (SP-KV), a learned \emph{sparse-write} mechanism that selectively writes only the most useful key--value pairs into the persistent KV cache. A lightweight utility predictor assigns a utility score to each KV pair; recent tokens remain accessible through a local causal window, while older KV pairs are written to and attended from the global cache only when their predicted utility exceeds a threshold. The language model and utility predictor are trained at scale jointly using only next-token prediction, typically through continual pretraining from a pretrained full attention checkpoint. Across model scales and sequence distributions, SP-KV enables high KV-cache reductions leading to memory and decoding speed improvements, with negligible degradation in validation loss or downstream evaluations. We provide extensive ablations and show that the mechanism transfers beyond dense attention to hybrid local-global settings.
\end{minipage}
}

\fcolorbox{gray!10}{gray!10}{
\begin{minipage}{0.975\linewidth}
\noindent\textbf{Contribution 2.} 
We further show that SP-KV can be used as an architectural probe to design stronger global/local attention transformer hybrids. Typically, by retaining only heads with the highest learned average SP-KV utility on a reference model as global, while making the remaining attention heads local, we obtain hybrids that outperform standard interleaved layouts at the same KV-cache budget. 
\end{minipage}
}

\section{Method Overview}

\subsection{Self-Pruned KV Mechanism}
\label{sec:gated-kv}

\autoref{fig:overview} illustrates the \emph{Self-Pruned KV} mechanism at layer $l$ for a single attention head.
Let $T$ be the sequence length and let $H^l = [h_0^l,\dots,h_{T-1}^l]^\top \in \mathbb{R}^{T \times d_{\text{model}}}$ denote the hidden states at layer $l$.

\paragraph{Key-wise utility prediction.}
For each key head $k$, we predict a utility value at each token position $s$:
\begin{equation}
u_s^{l,k} = \sigma\!\big(f_{\theta}^{l,k}(h_s^l)\big) \in (0,1)
\end{equation}
where \(\sigma(\cdot)\) denotes the logistic sigmoid function, ensuring the utility lies in \((0,1)\) and \(f_{\theta}^{l,k}(\cdot)\) is a lightweight utility predictor parameterized by \(\theta\), 
 a 2 layer perceptron (MLP). To simplify notation in what follows, we suppress the layer and key-head superscripts, denoting $u_s^{l,k}$ simply as $u_s$.
 During inference, we threshold the utility gate value with a threshold $\tau$ to obtain a binary value; $z_s=1$ means the KV pair at position $s$ is eligible for \emph{long-range} (global) attention; $z_s=0$ means it is not.
\begin{equation}
z_s =  \mathbf{1}\!\left[u_s\ge \tau\right], \qquad z_s\in\{0,1\}.
\end{equation}

\paragraph{Sliding-window \& gated global attention.}
To preserve local temporal features, we \emph{always} allow attention within a causal local  sliding window of size $w$ (by default $128$). For query position $t$ and key position $s$, we define the window indicator
\[
\mathbf{1}_{\mathrm{win}}(t,s) \;=\; \mathbf{1}\left[0 \le t - s  < w \right].
\]
The keys outside the local window are accessible only if they are gated on ($z_s =1$). This yields the effective availability mask

\begin{equation}
g_{t,s}
=
\begin{cases}
0 & \text{if } \bigl(z_s=1 \;\vee\; \mathbf{1}_{\mathrm{win}}(t,s)\bigr), \\[4pt]
-\infty & \text{otherwise.}
\end{cases}
\end{equation}

Let $M_{\mathrm{causal}}(t,s)\in\{0,-\infty\}$ be the standard causal mask bias (0 if $s\le t$, $-\infty$ otherwise). We combine this causal mask and the gating into a single additive bias, as 
\begin{equation}
B_{t,s} \;=\;
M_{\mathrm{causal}}(t,s) \;+ 
g_{t,s}.  
\end{equation}

The resulting gated attention for query position $t$ is then given by
\begin{equation}
o_t
\;=\;
\sum_{s}
\mathrm{softmax}\!\left(
\frac{\langle q_t,k_s\rangle}{\sqrt d} + B_{t,s}
\right)\, v_s. 
\end{equation}

\subsection{Training}

\paragraph{Phase 1. Soft Gated Training.}
The thresholding operation would break the gradient flow during backpropagation. 
To preserve differentiability during training, we remove the thresholding operator and instead, we add to the attention mask bias the logarithm of the utility prediction itself (soft gating). 
At the extremes, a utility of 0 leads to a negative infinity mask value (like the causal mask), a utility of 1 leads to a mask bias of 0 that is inconsequential to the attention: 
\begin{equation}
\widetilde{g}_s
\;=\;
\log u_s . 
\end{equation}

While models can be trained from scratch with gating (see \autoref{sec:app_scratch}), we mostly focus on sparsifying pretrained dense models during a relatively short continued pre-training (or midtraining) phase to boost downstream efficiency. To smoothly transition from dense to SP-KV attention, all KV Utility gates are initialized with a large positive bias, rendering them fully open (utility of 1) once SP-KV training starts. The model learns to gate certain keys off (sparsification) during training without any dedicated loss: we only optimize the vanilla next token prediction loss.
An alternative training strategy with binary gates during training is described and evaluated in \autoref{sec:app_hard_gated_training} and relies on stochastic sampling and a straight-through estimator \citep{Bengio2013EstimatingOP}.

\paragraph{Phase 2. Threshold-Aware Hard Gating.} To reduce the test-time discrepancy, we finish training with a phase of Thresholding-Aware Hard Gating (TAHG). After training with soft gating for the first $75\%$ of the cosine-decay schedule, we  freeze the Utility Predictor weights and binarize the utility gate with threshold $\tau \in [0, 1]$. 
This preserves the optimization advantages of soft gating and bidirectional gradient flow early in training, while improving alignment with inference-time sparsification regimes.

To preserve the sharp regime change associated with gate binarization while avoiding optimization instability, we smooth the transition through annealing for models trained with 32k context windows. Concretely, over 500 steps, we linearly interpolate between continuous and binary gates:
\begin{equation}
\tilde{u} = (1-\alpha)\, u + \alpha \,\mathbf{1}[u \geq \tau],
\end{equation}
where $\alpha$ is ramped linearly from 0 to 1. As a result, both the attenuation of gates below threshold and the amplification of gates above threshold are introduced progressively.

\paragraph{Training Efficiency.}
We implement gated attention as a modification of Flash Attention~3~\citep{shah2024flashattention3} for Hopper GPUs. The gate bias $\log u_s^{l,h}$ is fused directly into the pre-softmax score accumulation within FA3's online softmax, adding negligible arithmetic cost per element. In the forward pass, gate values are prefetched into registers once per KV-block column, avoiding redundant reloads across query rows within each tile. In the backward pass, gate gradients $\partial\mathcal{L}/\partial \log u_s$ are accumulated via atomic additions, since multiple query positions contribute to each gate's gradient. During Phase~2 (TAHG), binarized gates enable a \emph{block-skipping} optimization: before the kernel launch, we precompute a per-head sparsity mask at 64-token granularity, marking blocks where all gates are zero. The kernel skips both TMA loads and MMA compute for any KV block that is simultaneously all-zero and entirely outside the sliding window for the current query tile. At high sparsity this recovers most of the baseline throughput.

Across four model sizes (1.6B--8.1B parameters), soft-gated training (Phase~1) reaches $60\text{--}62\%$ of the wall-clock throughput of a full-attention baseline using FlashAttention-3 kernels. In Phase~2 ($\tau{=}0.7$), the resulting sparsity enables block skipping and improves throughput to $82\text{--}90\%$ of baseline. Gains are most pronounced for smaller models, where attention represents a larger share of total compute.

\paragraph{Measuring sparsity.} With \(z_s^{l,k}\in\{0,1\}\) the thresholded key gate for token position \(s\), layer \(l\), and key head \(k\), we define the \emph{gate density} as $\rho$ and refer to \(1-\rho\) as \emph{sparsity}. Note that a local window of size \(w\) is always retained; values reported only reflect the fraction of KV entries kept in the long term KV cache. $\rho$ is computed as 
\begin{equation}
\rho \;=\; \frac{1}{LKT}\sum_{l=0}^{L-1}\sum_{k=0}^{K-1}\sum_{s=0}^{T-1} z_s^{l,k}.
\end{equation}

\section{Experiments}
\label{sec:experiments}

Self-Pruned KV Attention has two design principles; reducing key-value cache memory size and improving decoding latencies while maintaining equivalent performance to full attention variants. Our experiments validate these principles through scaling analysis and a varied set of perplexity and downstream results, while uncovering techniques to trade off between both objectives.

\paragraph{Overall protocol.}
All experiments follow a single continued-training protocol. We first train full-attention models with a ratio of 140 training tokens per non-embedding parameter (TPP) using a warmup-stable learning rate schedule. Starting from this shared 140~TPP checkpoint, we branch training for an additional 20~TPP with a cosine decay scheduler into two models: (i) continuing with full attention (baseline), or (ii) switching the attention mechanism to \emph{Self-Pruned KV Attention}. This experimental design isolates the effect of the attention modification to the final stage of training while keeping the data, optimizer, and compute budgets matched. All models are trained on 8k sequence lengths, and 8.1B models are context extended during the last 10 TPP to 32k context lengths. Hyperparameters are rigorously optimized and choices are detailed in \autoref{sec:app_scaling}.

\begin{figure}[ht]
    \centering
     \includegraphics[width=0.75\linewidth]{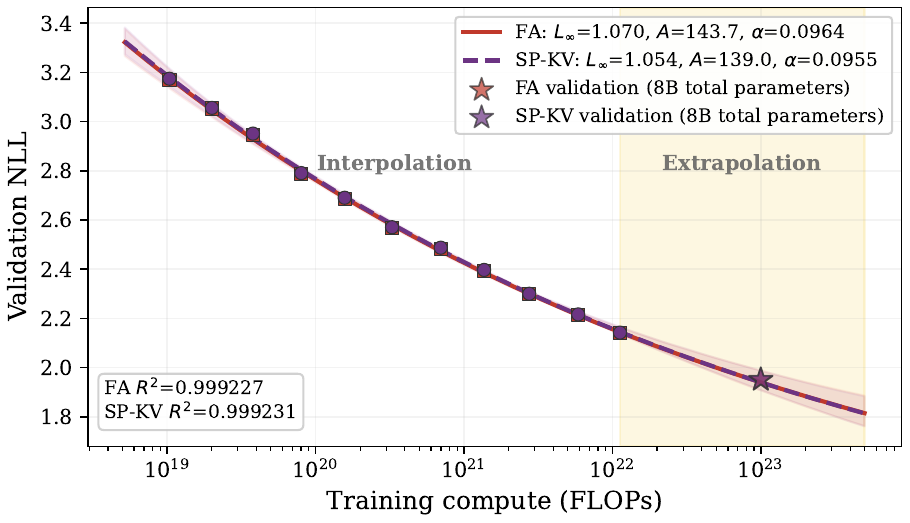}
    \caption{Validation NLL and training compute form an empirical power law. Fits for full attention models and Self-Pruned KV variants ($\tau=0.5$) are done across 11 different compute regimes.
    Both model families evolve closely with respect to compute. 8B models are not used for fits and empirically validate the extrapolation.}

    \label{fig:inf_scaling}
\end{figure}

\paragraph{Training details.} All models are pretrained on a standard pretraining data mixture composed primarily of DCLM data \citep{li2025datacomplmsearchgenerationtraining}, together with code, books, and several additional sources chosen in particular to increase the proportion of long sequences in the training corpus. Importantly, the training data contains only naturally occurring sequences. Performance on RULER should therefore be interpreted as fully out-of-distribution. We train Llama~3-based models with a fixed token-to-non-embedding-parameter ratio of 160, corresponding to approximately $8\times$ the compute-optimal ratio. We report validation negative log-likelihood (NLL) as the primary pretraining metric and evaluate downstream performance on a fixed task suite (full details are provided in \autoref{sec:app_additional_results}). 
All models rely on standard Grouped Query Attention (GQA) which already reduces the KV-cache size by 4-6$\times$ compared to MHA. 
Note that initial experiments using MHA resulted in much higher sparsification ratios, as keys are more redundant.

\subsection{Scaling analysis}
To validate SP-KV scales well with compute, we train a family of models spanning several orders of magnitude: $48$M to $7.0$B non-embedding parameters. Models are designed such that width-to-height ratio, attention query heads to key heads ratio, and FFN to hidden dimension ratio are approximately the same.
Hyperparameters are adjusted as a function of the training compute budget to ensure stable training across scales; complete configurations and heuristics are provided in \autoref{sec:app_scaling}.

Following \citet{deepseekai2024deepseekllmscalingopensource}, we fit a one-dimensional power law to the empirical negative log-likelihoods (NLLs) of models up to 2.24B parameters, evaluated on the LongPPL validation set \citep{fang2025wrongperplexitylongcontextlanguage}. LongPPL is chosen because longer sequences better reflect the long-context regime, although we observe consistent scaling trends across other validation splits. The law models NLL as a function of training compute $C$ in FLOPs. Because the parameter-to-token ratio is fixed in our setup, scaling is characterized by a single independent variable and three fitted parameters:
\begin{equation}
    L(C) = L_{\infty} + A C^{-\alpha}.
\end{equation}

\paragraph{Results.} The fits are highly accurate ($R^2 > 0.999$), see \autoref{fig:inf_scaling}. Both full attention and Self-Pruned KV Attention exhibit similar compute-scaling behavior. Extrapolations to larger compute budgets predict nearly identical performance for the two variants, and are confirmed by the 8.1B models unused for the fits. This indicates that adapting Self-Pruned KV via continual pretraining (1/8th of the full budget) does not degrade performance, while enabling the benefits of sparsity. Typically, only 29.6\% of the keys are kept for the 8.1B model ($\tau = 0.5$) on the validation data.

\subsection{Results on downstream tasks}

\begin{table}[t]
  \centering
  {\small
\begin{tabular}{llrrrr}
\toprule
Task & & Vanilla & Self-Pruned KV & $\Delta$ (\%) & Avg. Gate Density \\
\midrule
\multicolumn{6}{c}{\textbf{Short context evals }} \\[3pt]
ARC Challenge   & & 0.506 & 0.509 & \textcolor{ForestGreen}{+0.7\%} & 50.6\% \\
ARC Easy        & & 0.756 & 0.760 & \textcolor{ForestGreen}{+0.4\%} & 52.4\% \\
BoolQ           & & 0.704 & 0.730 & \textcolor{ForestGreen}{+3.7\%} & 25.2\% \\
CSQA            & & 0.675 & 0.676 & \textcolor{ForestGreen}{+0.1\%} & 28.0\% \\
GSM8k           & & 0.292 & 0.287 & \textcolor{BrickRed}{-1.6\%} & 18.7\% \\
Hellaswag       & & 0.785 & 0.784 & \textcolor{BrickRed}{-0.0\%} & 40.5\% \\
Human Eval Plus & & 0.280 & 0.262 & \textcolor{BrickRed}{-6.5\%} & 20.1\% \\
MBPP            & & 0.392 & 0.372 & \textcolor{BrickRed}{-5.1\%} & 24.8\% \\
MMLU            & & 0.559 & 0.560 & \textcolor{ForestGreen}{+0.1\%} & 28.4\% \\
NQ              & & 0.223 & 0.227 & \textcolor{ForestGreen}{+1.7\%} & 29.5\% \\
OBQA            & & 0.450 & 0.430 & \textcolor{BrickRed}{-4.4\%} & 49.9\% \\
PIQA            & & 0.806 & 0.803 & \textcolor{BrickRed}{-0.3\%} & 37.2\% \\
RACE High       & & 0.461 & 0.468 & \textcolor{ForestGreen}{+1.5\%} & 29.0\% \\
RACE Middle     & & 0.624 & 0.620 & \textcolor{BrickRed}{-0.6\%} & 28.9\% \\
TQA             & & 0.540 & 0.537 & \textcolor{BrickRed}{-0.5\%} & 29.2\% \\
Winogrande      & & 0.711 & 0.717 & \textcolor{ForestGreen}{+0.8\%} & 46.5\% \\
\midrule
\multicolumn{6}{c}{\textbf{Long context eval -- RULER}} \\[2pt]
\multirow{2}{*}{4k context}
& NIAH Single 1 & 1.000 & 1.000 & \textcolor{ForestGreen}{+0.0\%} & 6.7\% \\
& 4k Average (13 tasks) & 0.842 & 0.840 & \textcolor{BrickRed}{-0.2\%} & 18.8\% \\[4pt]
\multirow{2}{*}{8k context}
& NIAH Single 1 & 1.000 & 1.000 & \textcolor{ForestGreen}{+0.0\%} & 5.6\% \\
& 8k Average (13 tasks) & 0.793 & 0.786 & \textcolor{BrickRed}{-0.9\%} & 17.5\% \\[4pt]
\multirow{2}{*}{16k context}
& NIAH Single 1 & 1.000 & 1.000 & \textcolor{ForestGreen}{+0.0\%} & 5.2\% \\
& 16k Average (13 tasks) & 0.750 & 0.748 & \textcolor{BrickRed}{-0.3\%} & 17.0\% \\[4pt]
\multirow{2}{*}{32k context}
& NIAH Single 1 & 1.000 & 1.000 & \textcolor{ForestGreen}{+0.0\%} & 5.2\% \\
& 32k Average (13 tasks) & 0.635 & 0.610 & \textcolor{BrickRed}{-3.9\%} & 17.5\% \\[-4pt]
\multicolumn{6}{c}{\textcolor{black!30}{\rule{0.95\textwidth}{1pt}}} \\ 
Standard Average &  & 0.548 & 0.546 & \textcolor{BrickRed}{-0.2\%} & 33.7\% \\
RULER Average    &  & 0.755 & 0.746 & \textcolor{BrickRed}{-1.2\%} & 17.7\% \\
\bottomrule
\end{tabular}}
 \caption{Results on standard downstream tasks and the full RULER long-context benchmark (13 subtask types) for the 8.1B parameter model trained at 32k context with full attention, and its Self-Pruned KV variant ($\tau = 0.5$). SP-KV maintains standard benchmark performance ($-0.2\%$ average) while achieving ${\sim}66\%$ KV sparsification. 
 Overall RULER degradation is $-1.2\%$ with full per-task breakdown in \autoref{tab:ruler_full}. Density corresponds to the fraction of KV entries retained beyond the local window.}
 \label{tab:downstream_results_level13}
 \vspace{-0.5em}
\end{table}

Beyond perplexity, we test for non-regression w.r.t. to vanilla attention on a varied set of benchmarks.

\paragraph{Pretraining Benchmark Suite.} 
Using the 8.1B model trained at 32k context, we evaluate Self-Pruned KV attention on a broad suite of standard downstream benchmarks, reported in \autoref{tab:downstream_results_level13}. Overall, the Self-Pruned KV variant closely matches the full-attention baseline, with a negligible average change of $-0.2\%$ while retaining only $33.7\%$ of non-local KV entries on average. 

\paragraph{Long-Context Evaluation.} 
As shown in \autoref{tab:downstream_results_level13}, Self-Pruned KV preserves near-baseline performance on RULER benchmark tasks up to 16k tokens. The slightly larger drop at 32k ($-3.9\%$) likely reflects limited exposure to this regime, as 32k is the maximum context length seen during training. 
While these results demonstrate out-of-domain generalization to long-context sequences, additional experiments with RULER-style data added in the training mix (\autoref{tab:ruler_datamix_results}) show that SP-KV benefits substantially from long-context training, matching or outperforming vanilla-attention variants trained on the same data on most RULER tasks.
Full RULER results are in \autoref{tab:ruler_full}.

\paragraph{Sparsity.} 
The retained KV density varies substantially across tasks, as reported in \autoref{tab:downstream_results_level13}. Standard downstream tasks typically retain between $20\%$ and $50\%$ of non-local KV entries, with higher densities on very short tasks such as ARC, OBQA, and Winogrande, and lower densities on generative tasks such as GSM8k, HumanEval Plus, and MBPP. In contrast, RULER evaluations exhibit much lower densities, around $17$--$19\%$ on average, and Needle in a Haystack (NIAH) requires only about $5$--$7\%$ retained KV entries while maintaining perfect retrieval accuracy. This supports previous findings \citep{liu2023lostmiddlelanguagemodels} claiming task relevant information is sparse in long-context inputs, allowing the model to discard most past KV entries beyond the local window.

\begin{figure}[t]
    \centering
    \begin{subfigure}[b]{0.48\textwidth}
        \centering
        \includegraphics[width=\linewidth]{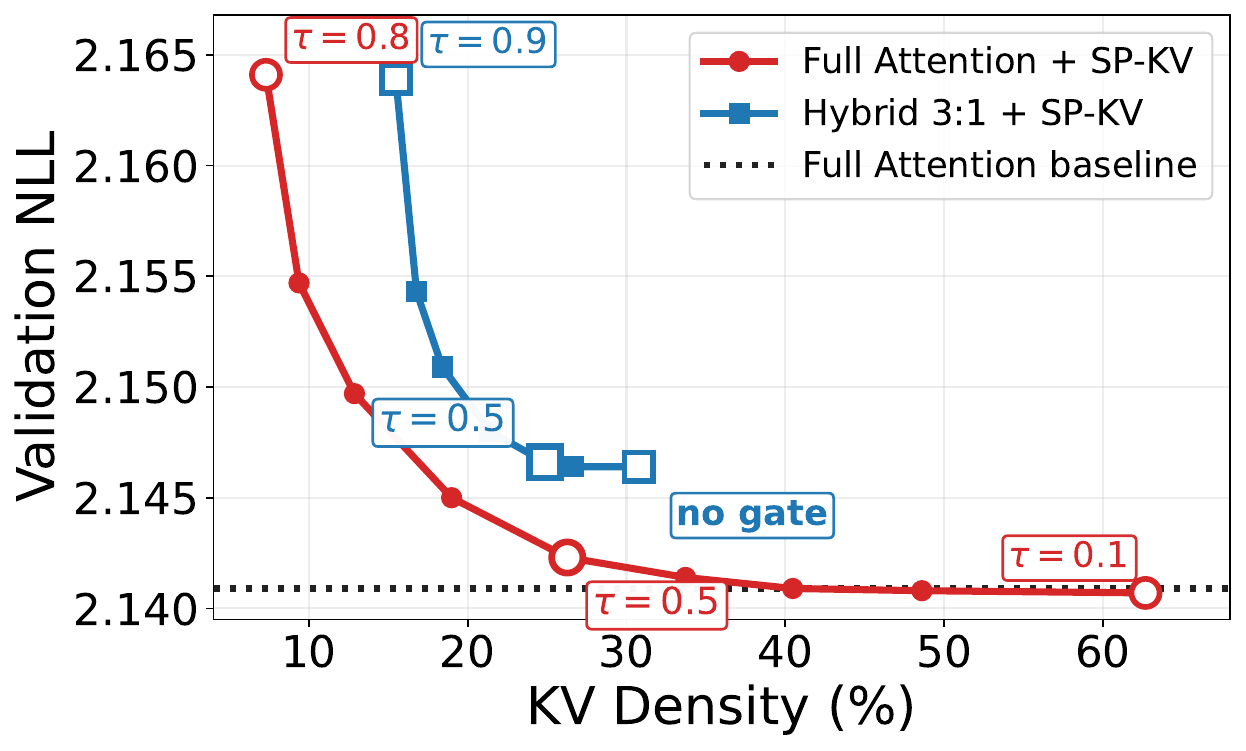}
    \end{subfigure}
    \hfill
    \begin{subfigure}[b]{0.47\textwidth}
        \centering
        \includegraphics[width=\linewidth]{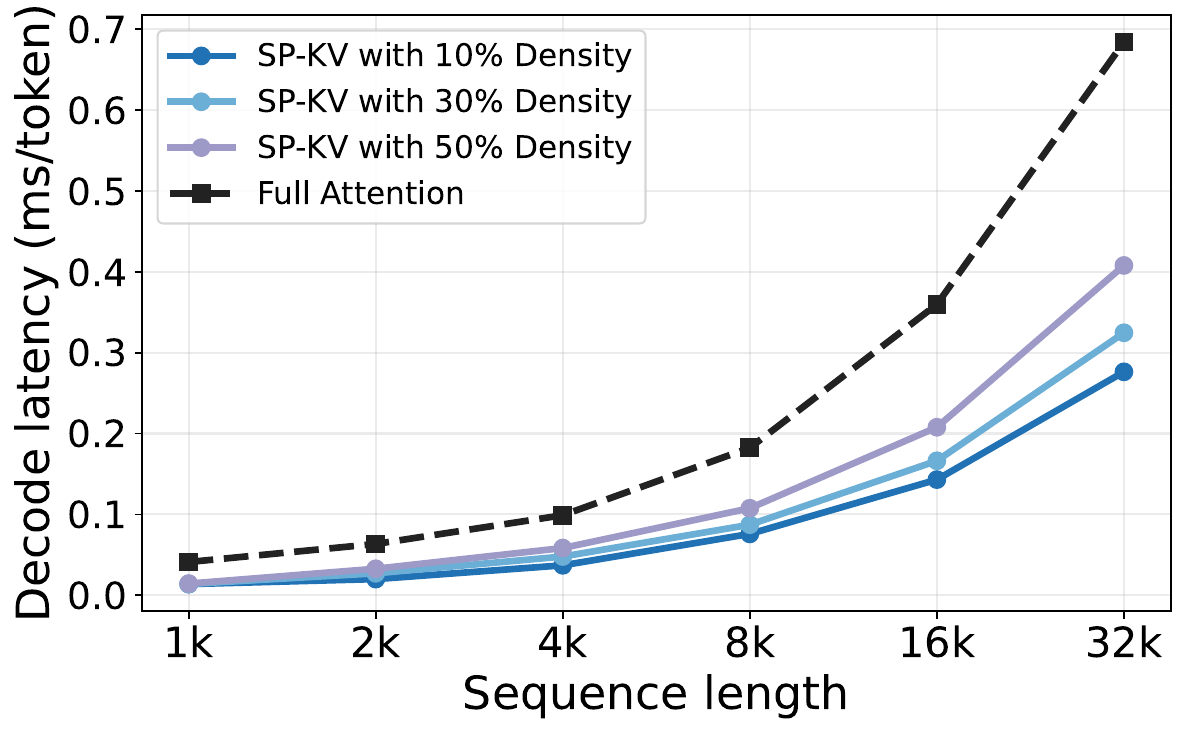}
    \end{subfigure}
    \caption{(Left) Relationship between NLL and KV cache density, when varying~$\tau$, the thresholding value that binarizes the gate decision. Experiments for a 2.91B model. Full attention + SP-KV Pareto-dominates the Hybrid 3:1 configuration \citep{faircodegenteam2025cwmopenweightsllmresearch}, achieving near-lossless NLL ({+0.07\%}) at ${\sim}26\%$ density ($\tau{=}0.5$). (Right) Per-token decoding latencies (ms) for a custom kernel implementation of Self-Pruned KV Attention (batch size 16). The memory bottleneck enables gated alternatives that limit key reads to outperform standard attention. Lower density ratios can directly translate to lower latencies.
\label{fig:latencies_decode}}
\end{figure}

\subsection{Controlling the Sparsity-Performance tradeoff}

Aggressively removing many KV entries reduces memory footprint and attention cost, but often leads to sharp quality degradation beyond moderate sparsity levels \citep{jegou2026kvzap}. In contrast, our method \emph{self-induces} sparsity through Self-Pruned KV attention training, yielding a substantially flatter degradation profile in practice.

\paragraph{Threshold Optimization.} The most direct sparsity  control is the pruning threshold $\tau$, which provides a smooth interpolation between retained KV density and downstream performance, as shown in \autoref{fig:latencies_decode} (left). Within the operating regime of SP-KV models (red), the model can achieve high sparsity, e.g., around $10\%$ retained density, while preserving strong performance; alternatively, using a denser setting recovers performance close to full attention.

\paragraph{Hybrids.} SP-KV can also be applied only to the global layers of transformers that interleave local and global attention (blue curve) \citep{riviere2024gemma2}. Such architectures are already sparse by design, since local layers do not maintain long-range KV caches, but their global layers can still be further sparsified with minor performance degradation up to a certain extent. Applying SP-KV to all layers yields a stronger sparsity--performance frontier as seen in \autoref{fig:latencies_decode}, but hybrid local/global variants are often easier to optimize for speed and memory during both training and inference.

Additional sparsity factors are studied in Subsection~\ref{sec:sparsity-vs-performance}. During training, these include the learning rate schedule, the utility predictor architecture, and the use of an auxiliary loss. At inference time, sparsity can be adjusted dynamically by changing the threshold value \textit{a posteriori}.

\subsection{Inference Efficiency}

Self-Pruned KV attention reduces the number of key--value entries stored and read during autoregressive decoding. This directly lowers KV-cache memory usage: at retained density $\rho$, the non-local cache footprint scales roughly as $\rho$ relative to full attention. As a result, SP-KV can support larger batches or longer contexts under the same memory budget.

We evaluate an initial sparse decoding kernel in \autoref{fig:latencies_decode} (right). The results show clear gains in memory-bound regimes, especially for batched long-context decoding. At batch size $16$, SP-KV kernels are consistently faster than full attention, with speedups of roughly $2.1\times$--$4.6\times$. In practice, gains shrink as density approaches $100\%$ and under shorter sequence lengths, since the SP-KV overhead offsets the reduction in KV reads. 
While further optimizations are possible, these results show SP-KV sparsity translates into practical efficiency gains through both reduced cache footprint and lower memory traffic during decoding.

\section{Designing Stronger Hybrids: SP-KV for Neural Architecture Search}

\paragraph{Using SP-KV as an architectural probe.} Beyond serving as an efficient attention mechanism, Self-Pruned KV Attention provides a direct signal about where long-range interactions are actually needed in a Transformer. Unlike post-hoc analyses of attention patterns in dense models, SP-KV induces a causal intervention: when a gate is closed, the corresponding key--value entry is unavailable to future tokens. Thus, retained gates identify interactions that the model learned to preserve because they matter for next-token prediction. 
Since the model trained with SP-KV  is not explicitly optimized to maximize sparsity, persistent high-density patterns reveal layers and heads that benefit from long-range access, while consistently sparse components suggest that local attention may suffice. We leverage this signal to study where global attention should be allocated in hybrid architectures.

\begin{figure}[t]
    \centering
    \begin{minipage}[t]{0.5\textwidth}
        \vspace{0pt}
        \centering
        \includegraphics[width=\textwidth]{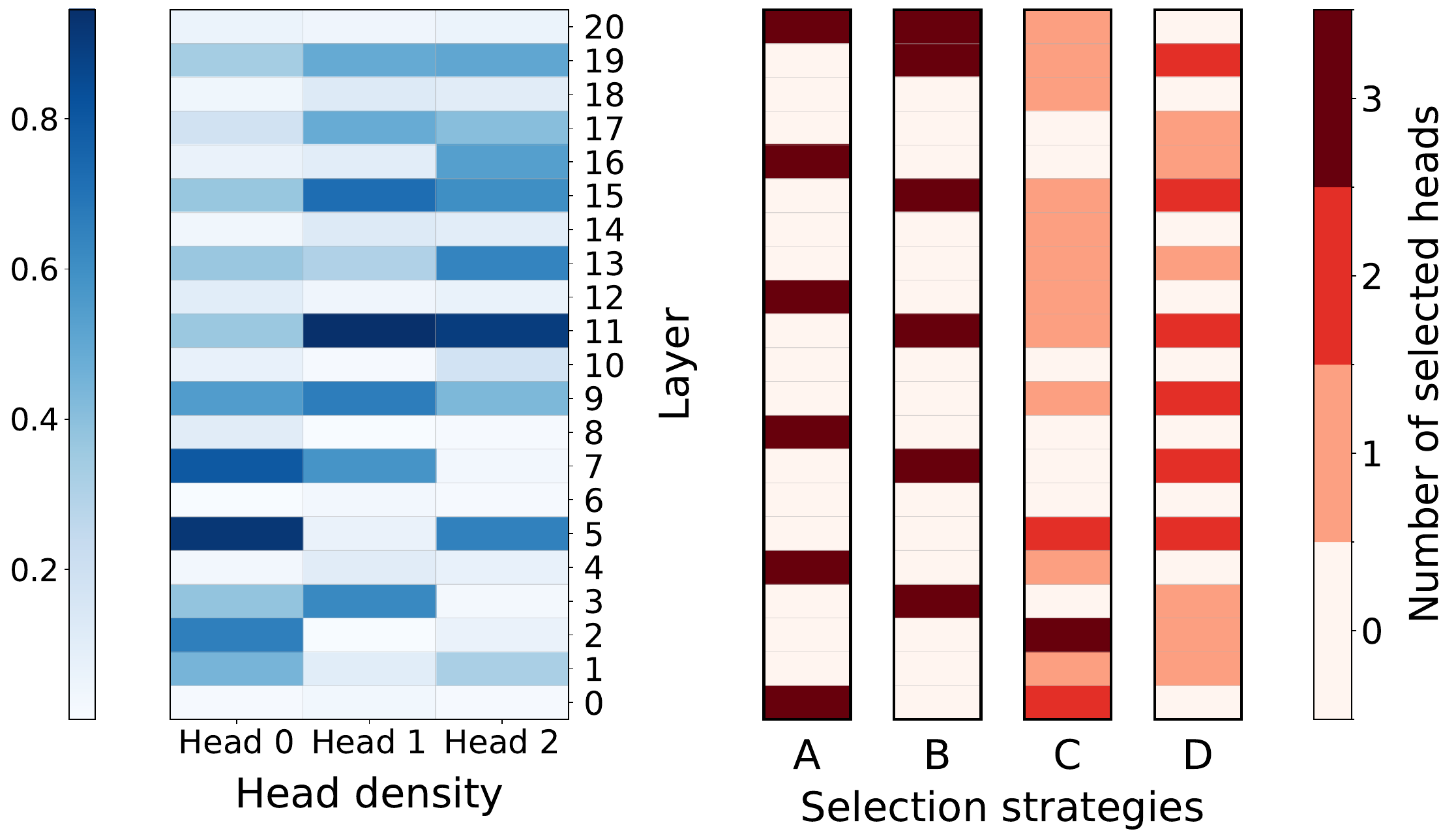}
        \label{fig:nas-strategies}
    \end{minipage}
    \hfill
    \begin{minipage}[t]{0.46\textwidth}
        \vspace{0pt}
        \centering
        \scriptsize
        \setlength{\tabcolsep}{3pt}
        \renewcommand{\arraystretch}{0.98}
        \begin{tabular}{@{}lcccc@{}}
        \toprule
        Head Selection Strategy & Density & Coverage & NLL & $\Delta$ NLL \\
                 & (\%) & (\%) & & (\%) \\
        \midrule
        \begin{tabular}[t]{@{}l@{}}
        A. 3:1 Layer Pattern \\
        \citepalias{faircodegenteam2025cwmopenweightsllmresearch}
        \end{tabular}
        & 28.6 & 8.28 & 2.3077 & \textcolor{BrickRed}{+0.396} \\
        
        \begin{tabular}[t]{@{}l@{}}
        B. 3:1 Layer Pattern \\
        \citepalias{cohere2025commanda}
        \end{tabular}
        & 28.6 & 45.65 & 2.3037 & \textcolor{BrickRed}{+0.222} \\
        
        C. 18 Random Heads
        & 28.6 & 34.34 & 2.3084 & \textcolor{BrickRed}{+0.426} \\
        
        D. 18 Densest Heads
        & 28.6 & 68.44 & \textbf{2.3023} & \textcolor{BrickRed}{+0.161} \\
        \midrule
        Full SP-KV ($\tau=0.5$)
        & 28.1 & 100.00 & \textbf{2.3003} & \textcolor{BrickRed}{+0.074} \\
        \midrule
        
        Full Global Attention
        & 100.0 & 100.00 & 2.2986 & - \\
        \bottomrule
        \end{tabular}
        \label{fig:nas-table}
    \end{minipage}
\vspace{-1em}
    
    \caption{Architecture search over sparse global-head layouts. \textbf{Left:} Learned per-head SP-KV density for a reference model informs four candidate 18-head selection strategies. \textbf{Right:} Each strategy’s selected density budget, density coverage, and downstream NLL, alongside full SP-KV and full global attention baselines.}
    \label{fig:nas}
\end{figure}

\paragraph{SP-KV-Guided Design Strategies.}
Recent hybrid architectures interleave local and global attention layers to reduce long context cost while preserving unrestricted token interactions in a subset of layers~\citep{riviere2024gemma2}. Our learned SP-KV gates provide a data-driven way to ask where these global interactions should be placed. The density patterns (representing the ratio of retained KVs) in \autoref{fig:nas} are highly non-uniform across layers and heads. By sorting heads by average SP-KV density and selecting only the top 18 ($28.6\%$ of heads) as global attention heads,  $68.4\%$ of the useful KV entries in the cache would be retained (\emph{density coverage}). 
Motivated by this observation, we compare several global-head allocation strategies in \autoref{fig:nas}. Strategies A and B follow fixed 3:1 local-global layer patterns inspired by prior hybrid architectures: Strategy A starts with a global layer at layer 0~\citepalias{faircodegenteam2025cwmopenweightsllmresearch}, whereas Strategy B delays the first global layer to layer 3, as in Command A~\citepalias{cohere2025commanda}. The last layer is global in both.
Strategy C relaxes the layer-wise constraint and selects 18 global heads uniformly at random. Finally, Strategy D uses SP-KV gate statistics from a reference model to optimally allocate the same budget of 18 global heads across layers to optimize \emph{density coverage}.

\paragraph{Results.} The
\emph{density coverage} (ratio of global keys from the reference model that would remain global under the new architecture) largely distinguishes the four architectures. Although all allocate the same global-attention budget (18/63 heads), they differ substantially in coverage. Strategy A only covers 8.28\% of the reference density patterns and performs poorly. Shifting the same 3:1 pattern to the Command-A-style offset (B) raises coverage to 45.65\%, which substantially reduces the degradation. 
Selecting the 18 densest heads (D) yields the highest coverage, 68.44\%, and achieves the best performance among fixed hybrid layouts. Logically, SP-KV at comparable density is a topline, with only $+0.074\%$ degradation, suggesting that learned sparsification within global attention provides additional flexibility beyond static head selection. However, fully local heads in hybrids are more straightforward to optimize during training, leading to reduced memory and compute costs.

\section{Related Work}
\label{sec:rw}

Prior work on KV-cache efficiency spans several complementary directions, including sparse retrieval, cache eviction or compression, quantization, and architectural constraints \citep{li2025surveykv}.  We distinguish \emph{sparse-read} methods, which reduce how much of the cache is accessed at inference time, from  \emph{sparse-write} methods, which reduce how tokens are retained in the cache in the first place.

\paragraph{Sparse-read methods.}
These methods reduce attention FLOPs or memory bandwidth by attending to only a subset of stored KV pairs while keeping the full cache available. They span query-aware token or block-level retrieval \citep{tang2024questqueryawaresparsityefficient,liu2024retrievalattentionacceleratinglongcontextllm}, landmark-based access patterns \citep{mohtashami2023landmarkattentionrandomaccessinfinite}, and external or reusable memory mechanisms such as kNN memories, block memories, or fast key retrieving subnetworks \citep{wu2022memorizingtransformers,xiao2024infllmtrainingfreelongcontextextrapolation,deepseekai2025deepseekv32pushingfrontieropen}. These approaches primarily target \emph{read-time} efficiency; they do not prevent the persistent KV state from growing with sequence length and are not direct comparisons to SP-KV. 

\paragraph{Sparse-write and cache-compression methods.}
Most related to our work, \textit {sparse-write} methods constrain KV growth by deciding which entries to retain, merge, or discard. Early eviction policies preserve recent tokens and attention sinks only (StreamingLLM, \citet{xiao2024efficientstreaminglanguagemodels}), or together with a small set of heavy hitters (H2O, \citet{zhang2023h2o}) or selected key tokens \citep{liu2023scissorhands,adnan2024keyformerkvcachereduction}. More recent methods improve token selection during prefilling or decoding using observation windows, head-aware scoring, or adaptive budget allocation
\citep{li2024snapkv,cai2024pyramidkv,feng2024adakv,tang2024razorattention}. Other approaches compress rather than simply evict entries, for example through inter-layer redundancy reduction or token merging \citep{liu2024minicachekvcachecompression,wang2024modeltellsmergeadaptive}. The best performing related recent direction learns utility estimates or eviction policies while operating on a frozen LLM, such as in ExpectedAttention~\citep{devoto2025expectedattentionkvcache}, KVZip~\citep{kim2025kvzip,kim2026fastkvzip}, and most recently KVZap~\citep{jegou2026kvzap}.
In these works, the sparsification mechanism is added \emph{post hoc} to a pretrained checkpoint, so the model is trained under dense attention and only sparsified at inference time. Conversely, DMS \citep{lancucki2025inferencetimehyperscalingkvcache} retrofits a pretrained LLM through short continued training with teacher logit distillation and sparsification objectives to learn an eviction policy. Further discussions on methods optimizing KV cache size through architectural constraints are discussed in Appendix \ref{app:kv_baseline_details}.

\begin{table}[t]
\centering
{\small \begin{tabular}{lrrr}
\toprule
Method & NLL & $\Delta$ NLL  & Avg Density \\
\midrule
StreamingLLM \citep{xiao2024efficientstreaminglanguagemodels} & 2.141 & +11.86\% & 0\% \\
StreamingLLM + 20\% Random & 1.992 & +4.09\% & 19.90\% \\
ExpectedAttention \citep{devoto2025expectedattentionkvcache} & 1.990 & +3.95\% & 20.70\% \\
$H_2$O \citep{zhang2023h2o} & 1.977 & +3.26\% & 20.45\% \\
KVZap \citep{jegou2026kvzap} & 1.991 & +4.01\% & 20.01\% \\
KVZap (+4 sink tokens) & 1.938 & +1.23\% & 20.15\% \\[-0.5em]

\multicolumn{4}{c}{\textcolor{black!30}{\rule{0.95\textwidth}{1pt}}} \\

StreamingLLM + 15\% Random & 2.009 & +4.98\% & 14.84\% \\
ExpectedAttention \citep{devoto2025expectedattentionkvcache} & 1.998 & +4.36\% & 16.44\% \\
$H_2$O \citep{zhang2023h2o} & 1.985 & +3.72\% & 15.74\% \\
KVZap (+4 sink tokens)  & 1.948 & +1.77\% & 15.15\% \\[-0.5em]

\multicolumn{4}{c}{\textcolor{black!30}{\rule{0.95\textwidth}{1pt}}} \\ 

(Ours) Self-Pruned KV Attention ($\tau = 0.5$) & 1.889 & \textbf{+0.08\%} & 25.72\% \\
(Ours) Self-Pruned KV Attention ($\tau = 0.7$) \hspace{2cm} & 1.896 & +0.46\% & \textbf{11.44\%} \\
\bottomrule
\end{tabular}}
\caption{Comparison of KV-cache reduction methods on the LongPPL validation set under a controlled evaluation protocol. Post-hoc baselines use the reference \texttt{kvpress} implementations with \textsc{Llama3.1 8B Instruct}, the checkpoint for which they were tuned or validated, while SP-KV is compared to its corresponding dense model trained with the same data and architecture. We report relative NLL degradation against dense baselines
and choose thresholds or compression ratios to match retained densities outside the local window. Sequence packing is disabled for fairness, so NLL values may differ slightly from the rest of the paper. See Appendix~\ref{sec:app_kv_baseline_details}.}
\label{tab:kv_baselines}
\vspace{-0.2em}
\end{table}

\paragraph{Baseline Comparisons.}
In contrast to these methods, SP-KV learns its utility predictor implicitly from the next token prediction objective, without auxiliary sparsification losses or direct utility supervision. This first enables joint optimization of the model weights and cache-selection policy at pretraining scale: in our 8B setting, SP-KV sees 140B tokens, nearly five orders of magnitude more than the calibration budget used for KVZap and about 50 times more than the retrofit budget reported for DMS-7B.
This distinction is reflected in Table~\ref{tab:kv_baselines}: at comparable retained densities, post-hoc baselines such as StreamingLLM, ExpectedAttention, and H$_2$O typically incur relative NLL increases of roughly $+3\%$ to $+5\%$, while KVZap is the strongest baseline in this family, reaching $+1.23\%$ at 20.15\% density and $+1.77\%$ at 15.15\% density when augmented with four sink tokens. SP-KV nevertheless yields the best trade-off overall, with only $+0.08\%$ relative NLL at 25.72\% density and $+0.46\%$ at 11.44\% density, supporting the view that exposing the model to sparsity during training reduces the train--test mismatch inherent to frozen-model pruning, and exposing head level granularity retains wide token coverage while enabling sparsity.
We detail the evaluation protocol in Appendix \ref{sec:app_kv_baseline_details}.

Importantly, unlike methods such as KVZap, which apply a single KV-cache compression step after prefill, SP-KV operates efficiently throughout decoding, enabling dynamic sparsification during autoregressive generation and thus sustained gains in memory usage and latency across the entire decoding process. Finally, because SP-KV is learned jointly with the language model rather than retrofitted afterward, it can be naturally carried into later training stages, such as instruction tuning or reinforcement learning, without requiring a separate policy-learning phase. This is particularly valuable for RL, where the mechanism could yield real efficiency gains during long rollouts, improving both training-time practicality and ease of use.

\section{Conclusion}

Self-Pruned KV attention is an effective mechanism to reduce the KV-cache bottleneck in LLMs. By learning which keys and values to write into persistent memory, SP-KV substantially lowers KV-cache size and improves decoding speed while maintaining strong performance across the settings we study. 
Beyond compression, SP-KV exposes informative sparsity patterns that identify where long-range token interactions matter most, enabling data-driven hybrid local/global architecture designs that improve upon standard interleaved baselines under the same KV-cache budget.

\paragraph{Future Work \& Limitations.} 
Experiments are conducted in a pretraining setting on a predominantly English-centric corpus, so it remains to be established how well the learned sparse-write policy transfers to multilingual settings, alternative data mixtures, or domains with different long-context statistics. Typically, our evaluation is centered on pretraining and standard downstream benchmarks. Studying behavior after post-training stages such as supervised fine-tuning and reinforcement learning, where the decoding distribution and the role of long-context memory may differ, would be an important next step.
Finally, although we report encouraging latency measurements, the systems implementation is not yet fully optimized. Converting KV sparsity into larger wall-clock gains will likely require dedicated kernels, improved scheduling, and memory layouts designed for sparse KV-cache access.

\bibliography{main}
\bibliographystyle{plainnat}

\appendix
\clearpage

~\vspace{3em}
\begin{center}
\Large{Appendices}
\end{center}
\vspace{3em}

\section{Additional Related Work and Details on Baseline Comparisons}
\label{app:kv_baseline_details}

\subsection{Further Linked Work}

\paragraph{Architecturally Constrained Cache Reduction.}
KV efficiency is often improved through mechanisms that are largely orthogonal to token selection, by reducing the cache size through architectural constraints. 
Grouped-query attention (GQA) shares key--value heads across multiple query heads, reducing the number of cached KV streams relative to MHA~\citep{ainslie2023gqa}; we use GQA throughout our experiments. 
More aggressive variants compress or share KV representations along other structural axes: Multi-head Latent Attention caches a low-dimensional latent state from which keys and values are reconstructed~\citep{liu2024deepseekv2}, while Cross-Layer Attention shares KV heads across adjacent layers~\citep{brandon2024reducingtransformerkeyvaluecache}, and quantization methods reduce the precision of cached states \citep{liu2024kivi}. 
Hybrid architectures instead constrain where global attention is available, typically by interleaving global layers with local sliding-window layers~\citep{beltagy2020longformer,riviere2024gemma2}, or by combining local attention with bounded-memory mechanisms such as compressive memories or recurrent sequence modules~\citep{munkhdalai2024leavecontextbehindefficient,yang2025gateddeltanet}. 
More recently, DeepSeek variants introduce fixed efficient memory hierarchies: DSA uses an indexer to retrieve only top-ranked KV entries~\citep{deepseekai2025deepseekv32pushingfrontieropen}, while DeepSeek-V4 introduces compressed attention mechanisms such as Compressed Sparse Attention, which compresses KV blocks before sparse retrieval, and Heavily Compressed Attention, which attends densely over a more aggressively compressed sequence~\citep{deepseekai2026deepseekv4}.

\paragraph{SP-KV is unconstrained.} These approaches improve efficiency by hard-coding a particular compression structure---which heads are shared, which layers are local, which latent state is cached, or how many keys can be accessed. 
SP-KV takes a different route: it keeps the full global cache as the candidate memory and learns which token-level key--value entries to write without constraints or external incentives to minimize cache size. 
Thus, SP-KV is not only complementary to architectural KV compression, but also strictly contains full attention as a limiting case: if all gates remain open, the mechanism recovers the original full-cache model. 
This flexibility matters in settings where fixed compression fails; our palindrome-reversal toy task in \autoref{app:palindrome_reversal} illustrates such a case, stressing the limitations of compression-based attention mechanisms while remaining solvable by a sparse-write policy that can retain the necessary tokens.

\subsection{Details on KV Compression Method Comparisons}
\label{sec:app_kv_baseline_details}

Table~\ref{tab:kv_baselines} compares SP-KV against representative post-hoc KV-cache compression methods in terms of the trade-off between retained cache density and relative NLL degradation. 
Because these methods are designed for different training regimes, absolute comparisons are irrelevant: most baselines operate on a frozen pretrained checkpoint, whereas SP-KV is trained jointly with the language model. 
We therefore take several steps to make the comparison as controlled and favorable to the baselines as possible.

\paragraph{Evaluation protocol.}
We use the reference implementations from \texttt{kvpress} and evaluate the baselines on Llama-3.1-8B-Instruct, the checkpoint for which these methods were tuned or validated. 
Our SP-KV models use the same Llama-style architecture, and all methods are evaluated on the same text sequences. 
Rather than comparing raw NLL values across different checkpoints, we report each method's NLL increase relative to its own dense baseline on the same examples, and compare this degradation against the average retained cache density outside the local window. 
This isolates the quality--compression frontier of each method while accounting for differences in pretraining, post-training, and model initialization.

\paragraph{Sparse decoding approximation.}
Many post-hoc KV-compression methods are implemented in a dense-prefill regime: the full prompt is first encoded with dense attention, and the cache is compressed only afterwards. 
Since our goal is to approximate incremental sparse decoding, we instead use a chunked-prefill protocol for perplexity evaluation. 
Prompts are processed in chunks of 16 tokens, and cache compression is applied after each chunk. 
All methods use an always-retained sliding window of size 128 and four sink tokens at the beginning of the sequence. 
Although sink tokens are not enabled by default for all reference implementations, we include them because they consistently improve baseline performance in this setting.

\paragraph{Post-hoc baseline performance.}
Among the post-hoc methods we evaluate, KVZap is the strongest baseline and thus the most informative point of comparison. 
In our setup, KVZap gives the best quality--density trade-off in this family, especially when augmented with four sink tokens. 
At approximately $20\%$ retained density, KVZap with sink tokens reaches $+1.23\%$ relative NLL, outperforming ExpectedAttention, H$_2$O, and random retention. 
At approximately $15\%$ density, it remains the best post-hoc baseline with $+1.77\%$ relative NLL. 
SP-KV substantially improves over this frontier, reaching $+0.46\%$ relative NLL at $11.44\%$ density and $+0.08\%$ at $25.72\%$ density.

\paragraph{Scope of the comparison.}
Efficient sparse-decoding kernels for many of these post-hoc baselines are not yet available. 
We therefore rely on the reference \texttt{kvpress} implementations and simulate pruning by masking or zeroing pruned keys so that they no longer contribute to attention. 
Consequently, Table~\ref{tab:kv_baselines} should be interpreted as a controlled quality comparison, not as a systems benchmark. 
In this form, several baselines are too slow to serve as practical sparse-decoding implementations.

\paragraph{Connection to long-context task performance.}
The NLL improvements in Table~\ref{tab:kv_baselines} are not merely a perplexity artifact. 
In \autoref{sec:app_additional_results}, we additionally train SP-KV with a small amount of RULER-style long-context data, in part to better match the post-trained regime of the evaluated benchmarks. 
As shown in \autoref{tab:ruler_datamix_results}, SP-KV improves average RULER-16k performance over the corresponding vanilla-attention model by $0.2\%$ while retaining only $15.3\%$ of keys. 
This is both sparser and stronger than the best reported KVZap setting, indicating that SP-KV can translate its compression advantage into downstream long-context performance when exposed to relevant training data.

\section{Hard Gated Training}
\label{sec:app_hard_gated_training}

This section complements \autoref{sec:gated-kv} by describing a strictly binary training variant of the gating mechanism.

\subsection{Binary gate sampling during training}
\label{app:hard-gate-sampling}

An alternative modeling to keep binary gates during training is hard gated training. To preserve gradient flow during training, we replace utility-value thresholding with stochastic sampling. Concretely, we sample a binary key gate
\begin{equation}
z_s^{l,k} \sim \mathrm{Bernoulli}(u_s^{l,k})  \qquad z_s^{l,k}\in\{0,1\}
\end{equation}
and
\[
g_{t,s}^{l,k}
=
\begin{cases}
0 & \text{if } \bigl(z_s^{l,k} =1 \;\vee\; \mathbf{1}_{\mathrm{win}}(t,s)\bigr),\\[4pt]
-\infty & \text{otherwise.}
\end{cases}
\]

\subsection{Straight-through gradient estimator}
\label{app:hard-gate-ste}

Because $z_s^{l,k}$ is discrete, we use a straight-through estimator (STE) to enable gradient flow into the utility predictor.
Operationally, we use the sampled log-gate in the forward pass but backpropagate as if the log-gate were $\log u_s^{l,k}$:
\begin{equation}
\widetilde{g}_s^{l,k}
\;=\;
\underbrace{g_s^{l,k}}_{\text{forward (no grad)}}\!\!. \texttt{detach()}
\;+\;
\log u_s^{l,k}
\;-\;
\big(\log u_s^{l,k}\big).\texttt{detach()}.
\end{equation}
Thus, the forward computation uses the hard gate (via the binary mapping from $z_s^{l,k}$), while the backward pass routes gradients through $\log u_s^{l,k}$.

\subsection{Results}

\begin{table}[t]
  \centering
  {\small 
\begin{tabular}{lrrrr}
\toprule
Task & Full Attention & SP-KV & $\Delta$ (\%) & Avg. Gate Density \\
\midrule
ARC Challenge & 0.493 & 0.505 & \textcolor{ForestGreen}{+2.4\%} & 31.3\% \\
ARC Easy & 0.780 & 0.770 & \textcolor{BrickRed}{-1.3\%} & 32.0\% \\
BoolQ & 0.729 & 0.681 & \textcolor{BrickRed}{-6.6\%} & 24.5\% \\
CSQA & 0.690 & 0.648 & \textcolor{BrickRed}{-6.1\%} & 22.7\% \\
GSM8k & 0.284 & 0.234 & \textcolor{BrickRed}{-17.6\%} & 17.4\% \\
Hellaswag & 0.785 & 0.785 & \textcolor{ForestGreen}{+0.0\%} & 28.0\% \\
Human Eval Plus & 0.226 & 0.244 & \textcolor{ForestGreen}{+8.0\%} & 18.5\% \\
MBPP & 0.368 & 0.386 & \textcolor{ForestGreen}{+4.9\%} & 21.0\% \\
MMLU & 0.564 & 0.544 & \textcolor{BrickRed}{-3.5\%} & 22.4\% \\
NQ & 0.222 & 0.210 & \textcolor{BrickRed}{-5.4\%} & 21.4\% \\
OBQA & 0.452 & 0.446 & \textcolor{BrickRed}{-1.3\%} & 34.1\% \\
PIQA & 0.792 & 0.792 & \textcolor{ForestGreen}{+0.0\%} & 29.8\% \\
RACE High & 0.459 & 0.450 & \textcolor{BrickRed}{-2.0\%} & 22.6\% \\
RACE Middle & 0.609 & 0.613 & \textcolor{ForestGreen}{+0.7\%} & 23.0\% \\
TQA & 0.530 & 0.528 & \textcolor{BrickRed}{-0.4\%} & 22.1\% \\
Winogrande & 0.717 & 0.722 & \textcolor{ForestGreen}{+0.7\%} & 30.8\% \\
\midrule
NIAH MultiKey 2 4096 & 0.964 & 0.942 & \textcolor{BrickRed}{-2.3\%} & 18.2\% \\
NIAH MultiKey 2 8192 & 0.856 & 0.670 & \textcolor{BrickRed}{-21.7\%} & 17.9\% \\
NIAH MultiQuery 4096 & 0.984 & 0.968 & \textcolor{BrickRed}{-1.6\%} & 18.9\% \\
NIAH MultiQuery 8192 & 0.953 & 0.917 & \textcolor{BrickRed}{-3.8\%} & 18.0\% \\
Variable Tracking 4096 & 0.958 & 0.965 & \textcolor{ForestGreen}{+0.7\%} & 11.1\% \\
Variable Tracking 8192 & 0.808 & 0.854 & \textcolor{ForestGreen}{+5.7\%} & 10.1\% \\
\midrule
Standard Average & 0.544 & 0.535 & \textcolor{BrickRed}{-1.7\%} & 25.1\% \\
RULER Subset Average & 0.920 & 0.886 & \textcolor{BrickRed}{-3.8\%} & 15.7\% \\
\bottomrule
\end{tabular}}
\smallskip
 \caption{Results on a suite of  standard downstream tasks for the  7.0B non-embedding parameter model (Llama3 8B) trained with full attention, and its equivalent Hard Gated Self-Pruned KV variant specialized with the mechanism for the last 1/8th of training. Self-Pruned KV maintains performance while enabling high levels of KV sparsification.}
 \label{tab:downstream_results_level13_hard}
\end{table}

\autoref{tab:downstream_results_level13_hard} shows hard gated evaluation results on the largest model size evaluated (7.0B non-embedding parameters). 
The hard gated training performs slightly worse than it's soft-gated equivalent, supposedly due to the discrepancy introduced by the STE.
Additional ablations on hard-gated variants are detailed in \autoref{app:ablation_details}.

\section{Extended Ablations and Sparsity Controls}
\label{app:ablation_details}

This section expands on the sparsity--quality trade-off analysis from Section~\ref{sec:experiments} by documenting additional ablations and control knobs.

\subsection{Ablating design choices}
\label{app:ablation-design}

Hard gated training results show very similar patterns to soft gating, with very slightly degraded performance. While this variant is conceptually more involved, it trains a little faster since we can leverage training sparsity in our custom attention kernels, and it provides more decisive gating decisions that help for Neural Architecture Search. We further ablate several design decisions that are key in our final soft-gated design.

\begin{table}[h]
\centering
\footnotesize
\begin{tabular}{l@{\hspace{-1em}}c@{\ \ }ccc}
\toprule
Model & Gate Density (\%) & $\Delta$ Density (\%) & NLL Token & $\Delta$ NLL (\%) \\
\midrule
Bernoulli Clipping $p$=0.01 & 33.0\% & \textcolor{BrickRed}{+30.0\%} & 2.316 & \textcolor{BrickRed}{+0.1\%} \\
Bernoulli Clipping $p$=0.05 & 37.0\% & \textcolor{BrickRed}{+45.9\%} & 2.324 & \textcolor{BrickRed}{+0.4\%} \\
Bernoulli Clipping $p$=0.1 & 38.6\% & \textcolor{BrickRed}{+52.2\%} & 2.330 & \textcolor{BrickRed}{+0.7\%} \\
\midrule
Local Window Size 1 & 60.7\% & \textcolor{BrickRed}{+139.3\%} & 2.352 & \textcolor{BrickRed}{+1.6\%} \\
Local Window Size 8 & 47.8\% & \textcolor{BrickRed}{+88.3\%} & 2.332 & \textcolor{BrickRed}{+0.8\%} \\
Local Window Size 32 & 33.4\% & \textcolor{BrickRed}{+31.9\%} & 2.321 & \textcolor{BrickRed}{+0.3\%} \\
Local Window Size 512 & 27.5\% & \textcolor{BrickRed}{+8.3\%} & 2.306 & \textcolor{ForestGreen}{-0.3\%} \\
Fixed Attention Sinks & 21.3\% & \textcolor{ForestGreen}{-16.0\%} & 2.312 & \textcolor{ForestGreen}{-0.1\%} \\
\midrule
LR Multiplier 0.1 & 82.7\% & \textcolor{BrickRed}{+226.1\%} & 2.302 & \textcolor{ForestGreen}{-0.5\%} \\
LR Multiplier 1 & 37.8\% & \textcolor{BrickRed}{+49.2\%} & 2.311 & \textcolor{ForestGreen}{-0.1\%} \\
\midrule
Soft Gating ( $\tau$=0.5 ) & 31.3\% & \textcolor{BrickRed}{+23.5\%} & 2.305 & \textcolor{ForestGreen}{-0.4\%} \\
Thresholding Aware Soft Gating  $\tau$=0.3 & 49.6\% & \textcolor{BrickRed}{+95.7\%} & 2.299 & \textcolor{ForestGreen}{-0.7\%} \\
Thresholding Aware Soft Gating  $\tau$=0.5 & 34.6\% & \textcolor{BrickRed}{+36.4\%} & 2.300 & \textcolor{ForestGreen}{-0.6\%} \\
Thresholding Aware Soft Gating  $\tau$=0.7 & 19.7\% & \textcolor{ForestGreen}{-22.2\%} & 2.308 & \textcolor{ForestGreen}{-0.3\%} \\
Thresholding Aware Soft Gating  $\tau$=0.9 & 2.6\% & \textcolor{ForestGreen}{-89.8\%} & 2.379 & \textcolor{BrickRed}{+2.8\%} \\
\midrule
Frozen LLM & 81.8\% & \textcolor{BrickRed}{+222.7\%} & 2.303 & \textcolor{ForestGreen}{-0.5\%} \\
Linear Utility Predictor & 33.3\% & \textcolor{BrickRed}{+31.4\%} & 2.312 & \textcolor{ForestGreen}{-0.1\%} \\
Utility Predictor Bias = 1 & 24.7\% & \textcolor{ForestGreen}{-2.7\%} & 2.315 & \textcolor{BrickRed}{+0.0\%} \\
Utility Predictor Bias = 20 & 42.4\% & \textcolor{BrickRed}{+67.0\%} & 2.311 & \textcolor{ForestGreen}{-0.1\%} \\
No Weight Decay in Utility Predictor & 25.7\% & \textcolor{BrickRed}{+1.1\%} & 2.314 & \textcolor{BrickRed}{+0.0\%} \\
Bidirectional Utility Predictor & 45.3\% & \textcolor{BrickRed}{+78.8\%} & 2.313 & \textcolor{ForestGreen}{-0.0\%} \\
\midrule
From Scratch Training (Linear) & 23.3\% & \textcolor{ForestGreen}{-8.3\%} & 2.308 & \textcolor{ForestGreen}{-0.3\%} \\
From Scratch Training (MLP Utility Predictor) & 15.8\% & \textcolor{ForestGreen}{-37.8\%} & 2.317 & \textcolor{BrickRed}{+0.1\%} \\
\midrule
Thresholding Aware Hard Gating $\tau$=0.3  & 33.6\% & \textcolor{BrickRed}{+32.6\%} & 2.310 & \textcolor{ForestGreen}{-0.2\%} \\
Thresholding Aware Hard Gating $\tau$=0.5 & 23.7\% & \textcolor{ForestGreen}{-6.4\%} & 2.311 & \textcolor{ForestGreen}{-0.1\%}\\

Thresholding Aware Hard Gating $\tau$=0.7 & 19.1\% & \textcolor{ForestGreen}{-24.6\%} & 2.313 & \textcolor{ForestGreen}{-0.0\%} \\
Thresholding Aware Hard Gating $\tau$=0.9 & 8.3\% & \textcolor{ForestGreen}{-67.5\%} & 2.325 & \textcolor{BrickRed}{+0.5\%} \\
\midrule
3 SWA : 1 Full Attention & 28.6\% & \textcolor{BrickRed}{+12.7\%} & 2.308 & \textcolor{ForestGreen}{-0.3\%} \\
3 SWA : 1 Full Attn $\rightarrow$ TASG ($\tau$=0.7) & 17.2\% & \textcolor{ForestGreen}{-32.2\%} & 2.309 & \textcolor{ForestGreen}{-0.2\%} \\
3 SWA $\rightarrow$ TASG ($\tau$=0.7) : 1 Full Attn & 29.1\% & \textcolor{BrickRed}{+14.7\%} & 2.307 & \textcolor{ForestGreen}{-0.3\%} \\
3 SWA $\rightarrow$ TASG ($\tau$=0.7) : 1 Full Attn $\rightarrow$ TASG ($\tau$=0.7) & 19.7\% & \textcolor{ForestGreen}{-22.4\%} & 2.310 & \textcolor{ForestGreen}{-0.2\%} \\
\midrule
Full Attention Baseline & 100.0\% & \textcolor{BrickRed}{+294.3\%} & 2.299 & \textcolor{ForestGreen}{-0.7\%} \\
Self-Pruned KV Baseline & 25.4\% & -- & 2.314 & -- \\
\bottomrule
\end{tabular}
\smallskip
\caption{Ablation Results on LongPPL Validation set \citep{fang2025wrongperplexitylongcontextlanguage} for the Level 8 models (1.05B non-embedding parameters). The Self-Pruned KV baseline is trained with no Bernoulli clipping, a learning-rate multiplier of 5 in the utility predictor, a local window size of 128, a weight decay of 0.1 in the utility predictor, and an initial utility predictor bias of 5. Deltas are relative \% change compared to the Self-Pruned KV baseline. We show our default parameters provide a reasonable trade-off between sparsity and performance.}
\label{tab:ablations}
\end{table}

\subsection{Controlling sparsity at train time}
\label{sec:sparsity-vs-performance}

This subsection adds implementation details behind the sparsity controls summarized in the main experiments (Section~\ref{sec:experiments}).

\paragraph{Predictor.} Each key-value pair is assigned a scalar utility \(u_s^{l,k}\in(0,1)\) via a lightweight predictor \(f_{\theta}^{l,k}\).
Overall, 2-layer MLP predictors tend to learn sharper, more selective utilities than linear predictors, typically yielding higher sparsity for comparable quality (\autoref{tab:ablations}).

\paragraph{Hyperparameters.} Beyond the predictor depth, our method exposes several training-time knobs that shape the emergent sparsity profile. 
Notably, we can modulate:
\begin{itemize} 
\item the predictor initialization parameters;
\item a bias and a standard-deviation multiplier control the initial gate-open rate and how quickly sparsity develops\footnote{We initialize gates to be nearly always open, and ablations confirm that a bias of \(5\) (yielding \(\sigma(5)\approx 0.993\)) together with a predictor learning-rate multiplier of \(5\) yields the lowest training losses.}.
\item adapting the learning rates of the utility predictor through a multiplier of the global LR also modulates the rate at which utilities polarize, with higher LR leading to higher sparsities. 
\item varying the local window size \(w\in\{64,128,256\}\) has a large impact on sparsity, with larger windows that preserve more local context by construction empirically yielding \emph{higher sparsity} ratios. 
\end{itemize}
We study these effects in \autoref{tab:ablations}.

\begin{figure}[t]
    \centering
    \includegraphics[width=0.7\textwidth]{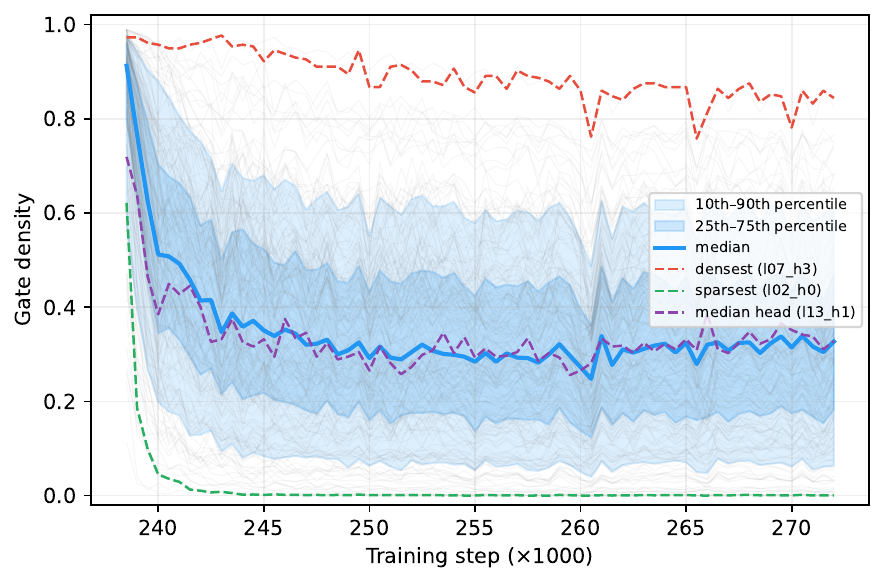}
    \caption{Per-head gate density during phase~1 soft gating CPT (FA, 8.1B).
    Solid blue: median across 190 heads. Shaded bands: 25th--75th and
    10th--90th percentiles. Dashed lines: representative individual heads
    (densest, sparsest, median). Thin grey: all 190 trajectories.}
    \label{fig:density-evolution}
\end{figure}

Figure~\ref{fig:density-evolution} illustrates this process on an 8.1B
full-attention model (L13, 190 heads). All gates initialize near 1.0
(all tokens retained). Within ${\sim}$12k steps of continued pretraining,
per-head densities diverge widely: some heads learn to retain most tokens
(density $\approx 0.84$) while others become highly selective
(density $\approx 0$). The median density settles around 0.33 with a
wide interquartile range (0.18--0.47), confirming substantial per-head
specialization without any explicit sparsity objective.

\paragraph{Bernoulli clipping.} We further consider \emph{Bernoulli probability clipping} during sampling to ensure the model observes a minimum rate of both open and closed gates:
\begin{equation}
\bar{u}_s^{l,k} \;=\; \mathrm{clip}\!\left(u_s^{l,k},\, p_{\min},\, 1-p_{\min}\right),
\qquad
z_s^{l,k} \sim \mathrm{Bernoulli}(\bar{u}_s^{l,k}).
\end{equation}
By preventing \(\bar{u}\) from saturating to \(0\) or \(1\) too early, this choice increases the frequency of ``rare'' gate events and (as we observe empirically) slows down sparsification while preserving stability.

 \paragraph{Regularization.}
  Although our main results rely on training \emph{without explicit sparsity incentives}, we find that an optional auxiliary loss can provide additional control over the final sparsity level.
  We adopt a simple \emph{density regularizer} that penalizes low mean utility:
  \begin{equation}
  \mathcal{L}_{\mathrm{aux}}
  \;=\;
  -\lambda_{\mathrm{aux}}\cdot
  \frac{1}{LKT}\sum_{l=0}^{L-1}\sum_{k=0}^{K-1}\sum_{s=0}^{T-1}
  u_s^{l,k}.
  \end{equation}
  Minimizing this term encourages utility values to remain high, slowing gate closure during training.
  By modulating $\lambda_{\mathrm{aux}}$, we can target a desired operating point along the sparsity--performance frontier: higher weights yield denser caches and improved validation loss, while lower weights permit more aggressive sparsification at the cost of quality degradation.
  As shown in \autoref{fig:init_thresh}, this mechanism enables smooth interpolation across a range of gate densities, complementing the inference-time threshold sweep and providing an additional axis for application-specific tuning.

\subsection{Controlling sparsity at inference time}
This subsection is the deployment-time counterpart of the training controls above, and directly supports the threshold sweeps reported in \autoref{sec:experiments}.

\begin{figure}[t]
    \centering
    \includegraphics[width=0.75\textwidth]{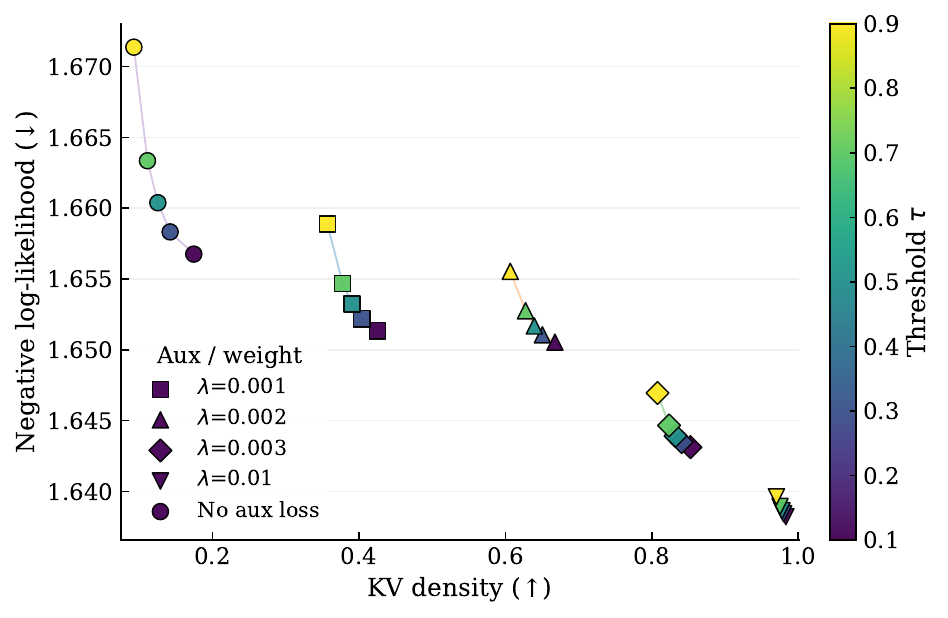}
    \caption{Model performance can be traded-off for additional sparsity by varying gate threshold values during inference, or by applying an auxiliary (aux) loss to regulate sparsification during training.}
    \label{fig:init_thresh}
    \end{figure}

At inference, we convert utilities into a deterministic keep/drop decision via a threshold \(\tau\):
\begin{equation}
\hat{z}_s^{l,k}(\tau) \;=\; \mathbf{1}\!\left[u_s^{l,k}\ge \tau\right],
\end{equation}
with \(\tau=0.5\) as the default.
Sweeping \(\tau\in[0.01,0.99]\) enables some control between performance and gate density, as reported in \autoref{fig:init_thresh} and \autoref{tab:ablations}. This threshold provides a simple post-training mechanism to dial compute/memory at deployment time without modifying model weights.

\subsection{Locally bidirectional utility predictor}
\label{sec:utility_noncausal}

We consider a variant that extends the utility-predictor design from \autoref{sec:gated-kv} by adding local look-ahead while preserving global causality.

Since we always retain a causal local window, the utility decision for a token only affects attention once that token falls outside the window. This delayed usage suggests an opportunity: we can condition the utility predictor on \emph{future} hidden states within the window without violating overall model causality.

We implement this by inserting a 1D convolution between the linear layers of the utility MLP. Using kernel size $65$ provides a $\pm 32$ token receptive field around each position, offering the predictor a short lookahead when deciding whether to keep a KV pair for long-range access. In practice, this modification slightly \emph{degrades} validation loss and downstream performance. We acknowledge more complex aggregation could be required to truly benefit from such a temporal aggregation, but as this also comes at the cost of a slight compute overhead, we leave this research direction for future work.

\subsection{Training from scratch}
\label{sec:app_scratch}

We have principally focused in this paper on continual pretraining setups and shown it was possible to sparsify a full attention model in later stages of training. However, we also experimented with training from scratch. Results exhibit similar patterns as those found in CPT setups, with full attention slightly outperforming Self-Pruned KV on validation loss. However, we notice greater levels of sparsity with only 15.8\% of gates retained on the 1.05B model, which is much sparser than the equivalent model trained with CPT. Results in \autoref{tab:ablations}.

\subsection{Frozen language model}
To assess the effect of joint training (as opposed to post-hoc training of the utility predictor, as in most prior work), we take the final checkpoint of the full-attention baseline, freeze the LLM weights, and train only the utility predictor for 20 TPP (approximately 10k steps for the 1.05B non-embedding model used in this ablation). While the NLL remains strong, sparsification under the same learning-rate conditions barely emerges: after training, the average gate density is above 80\%. These results highlight the benefit of allowing the model to jointly adapt to the sparsification mechanism.

\section{Additional results}
\label{sec:app_additional_results}

\begin{table}[h!]
  \centering
  {\fontsize{9.5pt}{9.5pt}\selectfont
  \setlength{\tabcolsep}{3.5pt}
  \renewcommand{\arraystretch}{0.94}
\begin{tabular}{lrrrr}
\toprule
Task & Vanilla & Self-Pruned KV & $\Delta$ (\%) & Avg. Gate Density \\
\midrule
\multicolumn{5}{l}{\textit{RULER --- 4k context}} \\
NIAH Single 1 & 1.000 & 1.000 & \textcolor{ForestGreen}{+0.0\%} & 6.7\% \\
NIAH Single 2 & 1.000 & 1.000 & \textcolor{ForestGreen}{+0.0\%} & 23.9\% \\
NIAH Single 3 & 0.984 & 0.988 & \textcolor{ForestGreen}{+0.4\%} & 23.1\% \\
NIAH MultiKey 1 & 0.992 & 0.914 & \textcolor{BrickRed}{-7.9\%} & 23.5\% \\
NIAH MultiKey 2 & 0.994 & 0.992 & \textcolor{BrickRed}{-0.2\%} & 16.5\% \\
NIAH MultiKey 3 & 0.976 & 0.978 & \textcolor{ForestGreen}{+0.2\%} & 16.4\% \\
NIAH MultiValue & 0.959 & 0.977 & \textcolor{ForestGreen}{+1.9\%} & 23.1\% \\
NIAH Multiquery & 0.991 & 0.992 & \textcolor{ForestGreen}{+0.1\%} & 23.8\% \\
Variable Tracking & 1.000 & 0.992 & \textcolor{BrickRed}{-0.7\%} & 8.3\% \\
CWE & 0.104 & 0.219 & \textcolor{ForestGreen}{+110.2\%} & 19.9\% \\
FWE & 0.873 & 0.825 & \textcolor{BrickRed}{-5.5\%} & 15.9\% \\
QA 1 & 0.582 & 0.566 & \textcolor{BrickRed}{-2.8\%} & 21.7\% \\
QA 2 & 0.492 & 0.482 & \textcolor{BrickRed}{-2.0\%} & 21.6\% \\
\midrule
\multicolumn{5}{l}{\textit{RULER --- 8k context}} \\
NIAH Single 1 & 1.000 & 1.000 & \textcolor{ForestGreen}{+0.0\%} & 5.6\% \\
NIAH Single 2 & 1.000 & 1.000 & \textcolor{ForestGreen}{+0.0\%} & 21.6\% \\
NIAH Single 3 & 0.862 & 0.954 & \textcolor{ForestGreen}{+10.7\%} & 21.6\% \\
NIAH MultiKey 1 & 0.972 & 0.920 & \textcolor{BrickRed}{-5.3\%} & 21.8\% \\
NIAH MultiKey 2 & 0.990 & 0.986 & \textcolor{BrickRed}{-0.4\%} & 15.6\% \\
NIAH MultiKey 3 & 0.818 & 0.792 & \textcolor{BrickRed}{-3.2\%} & 14.5\% \\
NIAH MultiValue & 0.897 & 0.952 & \textcolor{ForestGreen}{+6.0\%} & 21.7\% \\
NIAH Multiquery & 0.983 & 0.986 & \textcolor{ForestGreen}{+0.3\%} & 21.9\% \\
Variable Tracking & 0.998 & 0.996 & \textcolor{BrickRed}{-0.2\%} & 6.6\% \\
CWE & 0.016 & 0.005 & \textcolor{BrickRed}{-66.7\%} & 18.9\% \\
FWE & 0.813 & 0.734 & \textcolor{BrickRed}{-9.7\%} & 15.6\% \\
QA 1 & 0.471 & 0.420 & \textcolor{BrickRed}{-10.8\%} & 21.0\% \\
QA 2 & 0.486 & 0.470 & \textcolor{BrickRed}{-3.3\%} & 21.3\% \\
\midrule
\multicolumn{5}{l}{\textit{RULER --- 16k context}} \\
NIAH Single 1 & 1.000 & 1.000 & \textcolor{ForestGreen}{+0.0\%} & 5.2\% \\
NIAH Single 2 & 0.996 & 1.000 & \textcolor{ForestGreen}{+0.4\%} & 21.3\% \\
NIAH Single 3 & 0.756 & 0.812 & \textcolor{ForestGreen}{+7.4\%} & 21.3\% \\
NIAH MultiKey 1 & 0.954 & 0.902 & \textcolor{BrickRed}{-5.5\%} & 21.5\% \\
NIAH MultiKey 2 & 0.964 & 0.978 & \textcolor{ForestGreen}{+1.5\%} & 15.3\% \\
NIAH MultiKey 3 & 0.642 & 0.564 & \textcolor{BrickRed}{-12.1\%} & 13.8\% \\
NIAH MultiValue & 0.776 & 0.891 & \textcolor{ForestGreen}{+14.7\%} & 21.4\% \\
NIAH Multiquery & 0.929 & 0.933 & \textcolor{ForestGreen}{+0.4\%} & 21.5\% \\
Variable Tracking & 0.986 & 0.976 & \textcolor{BrickRed}{-0.9\%} & 5.8\% \\
CWE & 0.030 & 0.019 & \textcolor{BrickRed}{-37.1\%} & 17.3\% \\
FWE & 0.832 & 0.808 & \textcolor{BrickRed}{-2.9\%} & 16.0\% \\
QA 1 & 0.450 & 0.432 & \textcolor{BrickRed}{-3.9\%} & 20.1\% \\
QA 2 & 0.434 & 0.404 & \textcolor{BrickRed}{-6.9\%} & 20.9\% \\
\midrule
\multicolumn{5}{l}{\textit{RULER --- 32k context}} \\
NIAH Single 1 & 1.000 & 1.000 & \textcolor{ForestGreen}{+0.0\%} & 5.2\% \\
NIAH Single 2 & 0.996 & 0.998 & \textcolor{ForestGreen}{+0.2\%} & 22.3\% \\
NIAH Single 3 & 0.814 & 0.790 & \textcolor{BrickRed}{-2.9\%} & 22.3\% \\
NIAH MultiKey 1 & 0.848 & 0.800 & \textcolor{BrickRed}{-5.7\%} & 22.3\% \\
NIAH MultiKey 2 & 0.800 & 0.738 & \textcolor{BrickRed}{-7.8\%} & 15.6\% \\
NIAH MultiKey 3 & 0.354 & 0.294 & \textcolor{BrickRed}{-16.9\%} & 14.3\% \\
NIAH MultiValue & 0.553 & 0.506 & \textcolor{BrickRed}{-8.5\%} & 22.3\% \\
NIAH Multiquery & 0.763 & 0.656 & \textcolor{BrickRed}{-14.1\%} & 22.3\% \\
Variable Tracking & 0.885 & 0.876 & \textcolor{BrickRed}{-0.9\%} & 5.5\% \\
CWE & 0.009 & 0.006 & \textcolor{BrickRed}{-37.0\%} & 16.3\% \\
FWE & 0.610 & 0.656 & \textcolor{ForestGreen}{+7.6\%} & 17.1\% \\
QA 1 & 0.255 & 0.260 & \textcolor{ForestGreen}{+2.0\%} & 20.5\% \\
QA 2 & 0.362 & 0.346 & \textcolor{BrickRed}{-4.4\%} & 21.4\% \\
\midrule
RULER Average & 0.755 & 0.746 & \textcolor{BrickRed}{-1.2\%} & 17.7\% \\
\bottomrule
\end{tabular}}
\smallskip 
 \caption{Full RULER benchmark results (all 13 subtask types) for the 8.1B model trained at 32k context with Self-Pruned KV ($\tau = 0.5$). Results are grouped by evaluation sequence length. The main table (\autoref{tab:downstream_results_level13}) shows selected tasks and per-length averages. Degradation is concentrated in multi-needle retrieval tasks (NIAH MultiKey) and frequency extraction (FWE), while single-needle retrieval (NIAH Single 1/2) and variable tracking are largely preserved. Note that CWE baseline accuracy is very low ($<$10\%), making relative deltas unreliable for that task.}
 \label{tab:ruler_full}
\end{table}

\paragraph{Evaluation protocol.}
Short-context choice tasks use likelihood-based ranking
(the completion with the lowest NLL is selected):
ARC-C/E, BoolQ, HellaSwag, OBQA, PIQA, RACE-H/M, and Winogrande
are evaluated 0-shot; CSQA 7-shot; MMLU 5-shot.
Generation tasks use greedy decoding with exact-match scoring:
NQ and TriviaQA (5-shot), GSM8k (8-shot), HumanEval+ and MBPP
(3-shot, pass@1).
RULER~\citep{hsieh2024ruler} evaluates all 13 subtask types
(8 NIAH variants, Variable Tracking, CWE, FWE, QA$\times$2)
at sequence lengths 4K, 8K, 16K, and 32K with 500 samples each,
using greedy generation with task-specific string-match metrics.
Perplexity is measured over 200 prompts per domain across 9 validation domains.

\autoref{tab:ruler_full} presents all results aggregated in \autoref{tab:downstream_results_level13}. We notice RULER tasks do not uniformly degrade with respect to sequence length. Importantly, our pretraining mix does not contain any RULER-like data that would expose the SP-KV mechanism to the RULER task distribution. 

\paragraph{Long Context Training.}
We further test whether SP-KV benefits from in-distribution long-context supervision by adding a small fraction of RULER-style synthetic sequences to the training mix. For fairness, we retrain both SP-KV and the full-attention baseline on this datamix, with the same next-token prediction objective. In \autoref{tab:ruler_datamix_results}, both models improve substantially, and SP-KV nearly closes the gap to full attention, reducing the overall RULER degradation from $-1.2\%$ to $-0.3\%$. QA tasks, which are not included in the synthetic mix, also improve markedly, indicating that exposure to long-context structure transfers beyond simple task-specific pattern matching.

\begin{table}[h!]
  \centering
  {\fontsize{9.5pt}{9.5pt}\selectfont
  \setlength{\tabcolsep}{3.5pt}
  \renewcommand{\arraystretch}{0.94}
\begin{tabular}{lrrrr}
\toprule
Task & Vanilla & Self-Pruned KV & $\Delta$ (\%) & Avg. Gate Density \\
\midrule
\multicolumn{5}{c}{\textbf{Short context evals}} \\[3pt]
ARC Challenge & 0.489 & 0.510 & \textcolor{ForestGreen}{+4.2\%} & 50.1\% \\
ARC Easy & 0.752 & 0.766 & \textcolor{ForestGreen}{+1.8\%} & 51.8\% \\
BoolQ & 0.713 & 0.717 & \textcolor{ForestGreen}{+0.6\%} & 25.8\% \\
CSQA & 0.690 & 0.677 & \textcolor{BrickRed}{-1.9\%} & 28.9\% \\
GSM8k & 0.303 & 0.273 & \textcolor{BrickRed}{-9.8\%} & 18.7\% \\
Hellaswag & 0.782 & 0.784 & \textcolor{ForestGreen}{+0.2\%} & 40.5\% \\
Human Eval Plus & 0.287 & 0.250 & \textcolor{BrickRed}{-12.8\%} & 19.8\% \\
MBPP & 0.408 & 0.394 & \textcolor{BrickRed}{-3.4\%} & 24.9\% \\
MMLU & 0.560 & 0.563 & \textcolor{ForestGreen}{+0.5\%} & 28.3\% \\
NQ & 0.222 & 0.222 & \textcolor{ForestGreen}{+0.0\%} & 29.0\% \\
OBQA & 0.434 & 0.444 & \textcolor{ForestGreen}{+2.3\%} & 47.1\% \\
PIQA & 0.805 & 0.804 & \textcolor{BrickRed}{-0.1\%} & 34.2\% \\
RACE High & 0.467 & 0.470 & \textcolor{ForestGreen}{+0.6\%} & 29.2\% \\
RACE Middle & 0.621 & 0.625 & \textcolor{ForestGreen}{+0.7\%} & 29.2\% \\
TQA & 0.533 & 0.536 & \textcolor{ForestGreen}{+0.4\%} & 28.1\% \\
Winogrande & 0.713 & 0.708 & \textcolor{BrickRed}{-0.8\%} & 45.5\% \\
\midrule
\multicolumn{5}{c}{\textbf{Long context eval -- RULER}} \\[2pt]
\multicolumn{5}{l}{\textit{RULER --- 4k context}} \\
NIAH Average (8 tasks) & 1.000 & 1.000 & \textcolor{ForestGreen}{+0.0\%} & 16.1\% \\
Variable Tracking & 0.998 & 0.999 & \textcolor{ForestGreen}{+0.2\%} & 9.6\% \\
CWE & 0.938 & 0.912 & \textcolor{BrickRed}{-2.8\%} & 24.8\% \\
FWE & 0.892 & 0.878 & \textcolor{BrickRed}{-1.6\%} & 21.4\% \\
QA 1 & 0.531 & 0.558 & \textcolor{ForestGreen}{+5.1\%} & 21.9\% \\
QA 2 & 0.448 & 0.434 & \textcolor{BrickRed}{-3.1\%} & 21.9\% \\
4k Average (13 tasks) & 0.908 & 0.906 & \textcolor{BrickRed}{-0.2\%} & 17.6\% \\
\midrule
\multicolumn{5}{l}{\textit{RULER --- 8k context}} \\
NIAH Average (8 tasks) & 1.000 & 1.000 & \textcolor{ForestGreen}{+0.0\%} & 14.6\% \\
Variable Tracking & 1.000 & 1.000 & \textcolor{ForestGreen}{+0.0\%} & 7.6\% \\
CWE & 0.927 & 0.825 & \textcolor{BrickRed}{-11.0\%} & 22.7\% \\
FWE & 0.812 & 0.808 & \textcolor{BrickRed}{-0.5\%} & 20.3\% \\
QA 1 & 0.470 & 0.449 & \textcolor{BrickRed}{-4.4\%} & 21.1\% \\
QA 2 & 0.448 & 0.472 & \textcolor{ForestGreen}{+5.4\%} & 21.3\% \\
8k Average (13 tasks) & 0.897 & 0.889 & \textcolor{BrickRed}{-0.9\%} & 16.1\% \\
\midrule
\multicolumn{5}{l}{\textit{RULER --- 16k context}} \\
NIAH Average (8 tasks) & 1.000 & 0.999 & \textcolor{BrickRed}{-0.1\%} & 13.8\% \\
Variable Tracking & 1.000 & 1.000 & \textcolor{ForestGreen}{+0.0\%} & 6.7\% \\
CWE & 0.890 & 0.856 & \textcolor{BrickRed}{-3.8\%} & 21.6\% \\
FWE & 0.993 & 0.968 & \textcolor{BrickRed}{-2.5\%} & 19.7\% \\
QA 1 & 0.423 & 0.486 & \textcolor{ForestGreen}{+15.0\%} & 20.3\% \\
QA 2 & 0.386 & 0.412 & \textcolor{ForestGreen}{+6.7\%} & 20.8\% \\
16k Average (13 tasks) & 0.899 & 0.901 & \textcolor{ForestGreen}{+0.2\%} & 15.3\% \\
\midrule
\multicolumn{5}{l}{\textit{RULER --- 32k context}} \\
NIAH Average (8 tasks) & 0.994 & 0.994 & \textcolor{ForestGreen}{+0.0\%} & 13.9\% \\
Variable Tracking & 0.931 & 0.990 & \textcolor{ForestGreen}{+6.3\%} & 6.4\% \\
CWE & 0.827 & 0.684 & \textcolor{BrickRed}{-17.3\%} & 21.4\% \\
FWE & 0.976 & 0.942 & \textcolor{BrickRed}{-3.5\%} & 20.2\% \\
QA 1 & 0.260 & 0.310 & \textcolor{ForestGreen}{+19.4\%} & 20.6\% \\
QA 2 & 0.338 & 0.354 & \textcolor{ForestGreen}{+4.7\%} & 21.4\% \\
32k Average (13 tasks) & 0.868 & 0.864 & \textcolor{BrickRed}{-0.5\%} & 15.5\% \\
\midrule
Standard Average & 0.549 & 0.546 & \textcolor{BrickRed}{-0.4\%} & 33.2\% \\
RULER Average & 0.893 & 0.890 & \textcolor{BrickRed}{-0.3\%} & 16.1\% \\
\bottomrule
\end{tabular}}
 \caption{Results with RULER data mix: standard downstream tasks and full RULER benchmark (13 subtask types $\times$ 4 context lengths) for the 8.1B model at 32k context. Both models start from the same full-attention 32k checkpoint and are continued-pretrained during the cosine decay phase with ${\sim}1.7\%$ RULER synthetic data (11 of 13 task types, excluding QA) added to the standard pretraining mix. ``Vanilla'' denotes the full-attention variant; ``Self-Pruned KV'' adds annealed soft-to-hard gating ($\tau = 0.5$). Compared to the results without RULER data (\autoref{tab:ruler_full}), the data mix nearly eliminates the RULER degradation from gating (overall $\Delta$ from $-1.2\%$ to $-0.3\%$) while downstream performance remains unchanged ($-0.4\%$ average). CWE (common-words extraction) remains the primary source of gating cost, consistent with \autoref{tab:ruler_full}.}
 \label{tab:ruler_datamix_results}
\end{table}

\section{Inference efficiency}
\label{sec:app_inference_efficiency}

This section complements the inference-efficiency discussion in Section~\ref{sec:experiments} with compute accounting assumptions and kernel-level implementation details.

\paragraph{Inference compute.}
\citet{kaplan2020scalinglawsneurallanguage} approximate the forward-pass FLOPs for a decoder-only Transformer as a constant model size dependent term corresponding to linear layers plus an attention term that also depends on context length.
The formula is given as
\begin{equation}
C_{\mathrm{fwd/token}} \;\approx\; 2N \;+\; 2\,n_{\mathrm{layers}}\,n_{\mathrm{ctx}}\,d_{\mathrm{attn}},
\end{equation}
where $N$ is the number of (non-embedding) model parameters, $n_{\mathrm{layers}}$ is the number of Transformer layers, $n_{\mathrm{ctx}}$ is the context length (number of keys attended to), and $d_{\mathrm{attn}}$ is the total attention width.

\begin{figure}[h]
    \centering
    \includegraphics[width=0.99\linewidth]{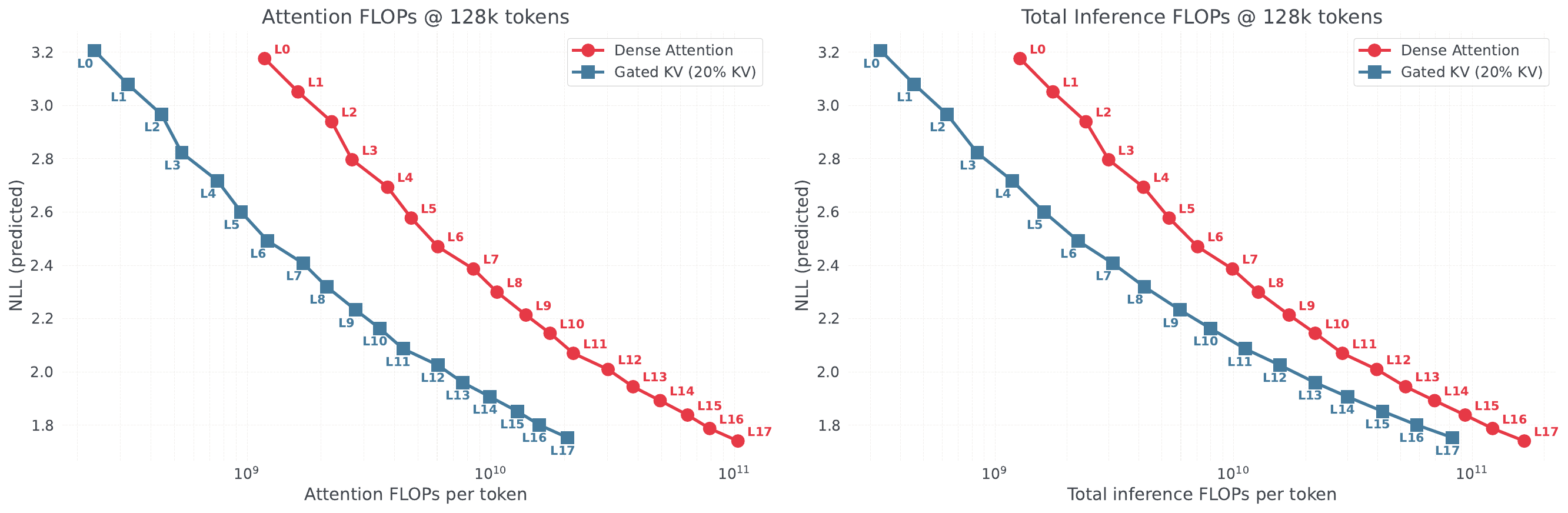}
    \caption{Per-token inference attention (left) and total (right) FLOPs for a cached context length of $L$ = 128k tokens. Using Kaplan-style accounting, dense attention follows $C(L)=2N+2\,n_{\mathrm{layers}}\,d_{\mathrm{attn}}\,L$, while $5\times$ sparsified KV (computing only $20\%$ of query--key interactions) follows $C_{\mathrm{sparse}}(L)=2N+0.2\cdot 2\,n_{\mathrm{layers}}\,d_{\mathrm{attn}}\,L$. Here $N$ is the non-embedding parameter count. In practice, latency is bottlenecked by cache read operations especially at longer sequence lengths, so cache size reductions directly translate in the same proportion to speed gains (most similarly to the left plot). }

    \label{fig:inf_compute}
\end{figure}

As $n_{\mathrm{ctx}}$ grows, the attention contribution increases linearly while the $2N$ term remains constant, so attention eventually dominates the per-token inference compute (Figure~\ref{fig:inf_compute}). Unlike commonplace KV-cache memory reduction techniques such as GQA or MLA, our sparsification method not only reduces the memory required to store the cache, but also avoids computing most query--key dot products. At the high sparsity levels achieved by our method (e.g., $80\%$ sparsity), the resulting FLOPs are equivalent to performing attention over a sequence that is $5\times$ shorter, effectively increasing the tractable context length by an order of magnitude. These compute gains are complementary to the memory savings, which also allow larger batch sizes by reducing KV-cache memory pressure.

\paragraph{Inference memory gains.}
At inference during decoding, the largest latency bottleneck lies not in the attention computation but rather in the KV cache read operations. As sequences grow longer, the total speed gains from reducing the size of the KV cache reads become asymptotically proportional to the size of the KV cache reduction in ideal settings.

\paragraph{Inference Kernels.} We implement inference decoding on a custom fork of FlashInfer~\citep{ye2024flashinfer}. Standard FlashInfer uses a tightly-packed paged KV cache: each request's page indices are stored contiguously via an \texttt{indptr} array with no slack, so appending a page to one request requires shifting all subsequent entries. SP-KV requires per-head variable-length caches (each head retains different tokens), making this tight packing impractical for autoregressive generation.

Our fork pre-allocates headroom in the page index array for each head, so that appending a page reduces to a single $O(1)$ \texttt{atomicAdd} on \texttt{used\_pages}[i] plus one index write. The capacity is grown dynamically when any head exhausts its headroom. All KV data, both the sliding-window region and retained long-range tokens, resides in a single paged pool; the window is tracked via lightweight metadata (a circular write pointer and per-slot gate decisions) rather than a separate buffer.

Token append during autoregressive decoding is handled by a single fused CUDA kernel per step. The kernel writes the new token into the next window slot; if the window is full, it checks the evicted token's gate: retained tokens are moved to the long-term region via atomic page allocation, while pruned tokens are discarded. Capacity checks (page pool and index expansion) are amortized, running only when a precomputed safe-step budget is exhausted.

While further work on inference efficiency is warranted to optimize the Self-Pruned KV mechanism, we show in \autoref{fig:inf_compute} that the theoretical inference efficiency gains enable running much larger models for equivalent inference compute. Assuming latency is memory-bottlenecked, the left subplot illustrating attention FLOPs closely resembles the expected gains for large sequence lengths.

\section{Scaling analysis details}
\label{sec:app_scaling}

We detail in \autoref{tab:scaling_nll} the model configurations used in our scaling laws.

\begin{table}[h]
  \centering
  {\small
  \begin{tabular}{
    l
    S[table-format=1.2]
    S[table-format=1.1e1]
    S[table-format=4.2]
    S[table-format=1.2]
  }
    \toprule
    {Scale} & {Non-Embed Params (B)} & {Train FLOPs} & {Total Tokens (B)} & {NLL} \\
    \midrule
    0  & 0.05 & 3.9e18 &  7.71  & 3.17 \\
    1  & 0.07 & 7.7e18 & 10.83  & 3.05 \\
    2  & 0.09 & 1.4e19 & 14.88  & 2.95 \\
    3  & 0.15 & 3.4e19 & 24.33  & 2.79 \\
    4  & 0.21 & 6.8e19 & 34.07  & 2.69 \\
    5  & 0.33 & 1.5e20 & 52.86  & 2.57 \\
    6  & 0.51 & 3.4e20 & 81.17  & 2.48 \\
    7  & 0.71 & 6.6e20 & 113.64 & 2.39 \\
    8  & 1.05 & 1.4e21 & 168.48 & 2.30 \\
    9  & 1.59 & 3.1e21 & 253.99 & 2.21 \\
    10 & 2.25 & 6.0e21 & 359.91 & 2.14 \\
    13 & 7.14 & 5.7e22 & 1141.68 & 1.95 \\
    Llama3 8B & 6.98 & 5.6e22 & 1116.73 & 1.95 \\
    \bottomrule
  \end{tabular}}
  \smallskip
    \caption{Main scaling study parameters per compute level: non-embedding parameters, training compute, total tokens, and negative log-likelihood (NLL) across model scales for the base model trained with WSD schedule. Importantly, total tokens are linked to non-embedding parameters by a ratio of 160:1, to replicate standard inference "optimal" use cases. The ladder is designed to regularly space training compute FLOPs.}
    \label{tab:scaling_nll}
\end{table}

Hyperparameters per compute level are obtained through hyperparameter scaling laws on lower compute levels \citep{deepseekai2024deepseekllmscalingopensource}. We obtain batch sizes and learning rates for each compute level detailed in \autoref{tab:scaling_params}.

\begin{table}[h]
  \centering
  {\small
    \begin{tabular}{l@{\quad\quad\quad}rrrrrrrrrr}
    \toprule
     & $d_{\text{model}}$ & $L$ & $H$ & $H_{\text{kv}}$ & Batch (tok) & $\eta$\ \ \ ~ & Steps & FLOPs & Warmup & Decay Start \\
    \midrule
    0 & 640 & 7 & 5 & 1 & 655k & 2.06e-3 & 12k & $3.93 \times 10^{18}$ & 73 & 10k \\
    1 & 768 & 8 & 6 & 1 & 721k & 1.87e-3 & 15k & $7.66 \times 10^{18}$ & 93 & 13k \\
    2 & 768 & 11 & 6 & 1 & 852k & 1.70e-3 & 17k & $1.45 \times 10^{19}$ & 109 & 15k \\
    3 & 1024 & 10 & 8 & 2 & 1.05M & 1.50e-3 & 23k & $3.45 \times 10^{19}$ & 145 & 20k \\
    4 & 1024 & 14 & 8 & 2 & 1.18M & 1.35e-3 & 29k & $6.75 \times 10^{19}$ & 180 & 25k \\
    5 & 1280 & 14 & 10 & 2 & 1.31M & 1.20e-3 & 40k & $1.51 \times 10^{20}$ & 252 & 35k \\
    6 & 1536 & 15 & 12 & 2 & 1.57M & 1.07e-3 & 52k & $3.39 \times 10^{20}$ & 322 & 45k \\
    7 & 1536 & 21 & 12 & 2 & 1.84M & 9.67e-4 & 62k & $6.64 \times 10^{20}$ & 387 & 54k \\
    8 & 1920 & 21 & 15 & 3 & 2.10M & 8.66e-4 & 80k & $1.40 \times 10^{21}$ & 502 & 70k \\
    9 & 2304 & 23 & 18 & 3 & 2.62M & 7.71e-4 & 97k & $3.08 \times 10^{21}$ & 605 & 85k \\
    10 & 2560 & 26 & 20 & 4 & 3.15M & 6.98e-4 & 114k & $6.04 \times 10^{21}$ & 715 & 100k \\
    13 & 3840 & 38 & 30 & 5 & 4.19M & 5.01e-4 & 272k & $5.71 \times 10^{22}$ & 1701 & 238k \\
    Llama3 8B & 4096 & 32 & 32 & 8 & 4.19M & 5.02e-4 & 266k & $5.58 \times 10^{22}$ & 2k & 233k \\
    \bottomrule
    \end{tabular}}
    \smallskip
    \caption{Detailed scaling study parameters and fitted hyperparameters per compute level. Models are designed to maintain an approximately constant width-to-height ratio, query-to-key head ratio, and FFN-to-attention-dim ratio. Models are trained for 160 TPP (tokens per NE parameters). Warmup is done for 1 TPP, the stable phase at peak LR lasts until 140 TPP, and a cosine decay is done from peak to 1\% of peak LR during the ultimate 20 TPP. Batch sizes per compute level are reused from the work in \cite{faircodegenteam2025cwmopenweightsllmresearch}}
\label{tab:scaling_params}
\end{table}

\section{Palindrome reversal with long instruction gap}
\label{app:palindrome_reversal}

We design a toy task to illustrate the adaptability of \emph{Self-Pruned KV attention} on settings where standard restricted-context mechanisms (e.g., sliding-window attention) are known to fail.

\paragraph{Task construction.}
Each example consists of (i) an input sequence of $N{=}32$ two-digit integers from $\{00,\dots,99\}$, represented as text tokens and separated by a single whitespace; (ii) a deliberately long natural-language instruction string (longer than 50 tokens) inserted to increase the token distance between the input and the output; and (iii) the target output, which is the \emph{reversed} input sequence, again formatted as two-digit integers separated by whitespace. Concretely, if the input sequence is
\[
57\ \ 12\ \ \dots\ \ 70,
\]
then the desired output is
\[
70\ \ \dots\ \ 12 \ \ 57.
\]
The model is trained \emph{only} on the cross-entropy loss over the output sequence tokens (i.e., losses are not applied to the input sequence nor the instruction tokens).

\paragraph{Models compared.}
We compare the following attention variants:
\begin{enumerate}
    \item \textbf{Full attention:} standard dense causal attention over the entire context.
    \item \textbf{Sliding-window attention:} causal attention restricted to a fixed window size of 32 tokens.
    \item \textbf{Self-Pruned KV attention (local window 32):} a Self-Pruned KV attention mechanism using a local attention window of 32 tokens, with gating enabling selective retention and retrieval of key--value pairs beyond the local neighborhood.
\end{enumerate}

\paragraph{Training protocol.}
All models are trained from scratch on randomly generated sequences. We generate sufficiently many unique sequences such that the training set effectively contains no repeats. We use a 2.25B non-embedding parameter model with a batch size of 64, a cosine scheduler with standard hyperparameters and observe learning dynamics over training steps.

\begin{figure}[h]
    \centering
    \includegraphics[width=0.95\linewidth]{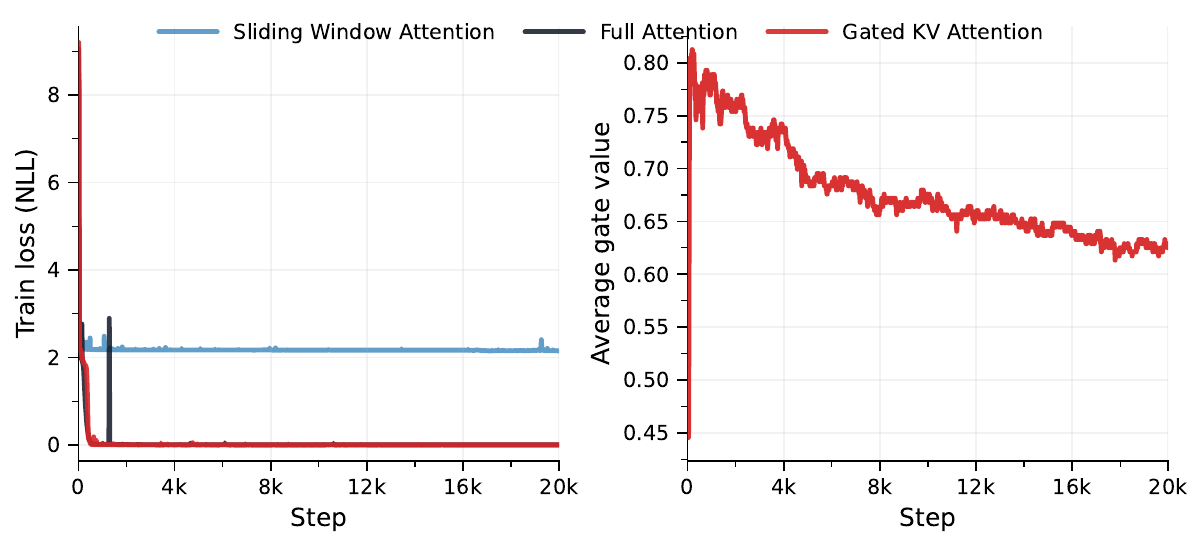}
    \caption{Palindrome reversal with a long instruction gap. Full attention and Self-Pruned KV attention (local window 32) rapidly converge to near-zero loss, while sliding-window attention (window 32) remains near chance. The right panel displays the mean utility values of KV gates.}
    \label{fig:palindrome_reversal_learning}
\end{figure}

\paragraph{Results.}
We find that both the full-attention model and the Self-Pruned KV attention model learn the task rapidly: after approximately 500 optimization steps (batch size 64), both reach near-zero loss on the output tokens. In contrast, the sliding-window attention baseline with window size 32 fails to solve the task and remains near chance-level performance, consistent with the method's inability to attend to KV from the input sequence during output sequence generation. A rough reference point for chance-level behavior is on the order of a NLL of 2.3, corresponding to $\ln(100)$ per two-digit number and perfect reconstruction of interleaved whitespace tokens.

\paragraph{Sparsification behavior.}
During training, Self-Pruned KV attention exhibits \emph{natural sparsification} of the KV cache: the gating mechanism increasingly concentrates the computation on a subset of relevant key--value pairs while still enabling the long-range dependencies required for perfect reversal. \autoref{fig:palindrome_reversal_learning} illustrates the training loss trajectories and the average utility prediction value over optimization steps.

These results suggest that, although Self-Pruned KV attention sparsifies the KV cache, it can do so \emph{without sacrificing} the computational expressivity of full attention on this class of long-range dependency tasks. In particular, gating does not inherently restrict the set of algorithmic sequence transformations the model can learn, even when the relevant input evidence is separated from the target output by a substantial token-distance gap.

\section{Compute Resources and Software}
\label{sec:app_compute_resources}

All training and evaluation runs reported in this work were conducted on NVIDIA H100 Hopper 80GB GPUs.  
Unless noted otherwise, experiments used the latest stable PyTorch release available in our environment at run time.

\section{Reference Implementation}
\label{sec:app_reference_impl}

\begin{lstlisting}[caption={SP-KV Attention reference implementation.},label={lst:spkv},float=t]
import torch, torch.nn as nn, torch.nn.functional as F

class UtilityPredictor(nn.Module):
    """Per-key, per-head utility predictor: u_s in (0, 1)."""
    def __init__(self, d_model: int, hidden: int, n_kv_heads: int):
        super().__init__()
        self.net = nn.Sequential(
            nn.Linear(d_model, hidden), nn.SiLU(),
            nn.Linear(hidden, n_kv_heads))

    def forward(self, h):                      # h: [B, T, d_model]
        return torch.sigmoid(self.net(h))      #  -> [B, T, n_kv_heads]
        # GQA: one gate per KV head; broadcast to query groups.

def sp_kv_attention(
    q, k, v,              # [B, H, T, D]
    utility,              # [B, H, T] in (0, 1)
    window_size=128,
    hard=False,           # True: binary threshold (inference)
    tau=0.5,
):
    """Causal attention with sliding window + SP-KV gating.

    Soft (training):  gate bias = log(u)           (differentiable)
    Hard (inference): gate bias = 0 if u>=tau else -inf  (binary)
    Within the window, all tokens attend regardless of gate.
    """
    B, H, T, D = q.shape

    if hard:
        gate = torch.where(utility >= tau, 0.0, float("-inf"))
    else:
        gate = torch.log(utility + 1e-8)       # [B, H, T]
        # Epsilon avoids NaN gradients if sigmoid saturates to exact zero.

    qi = torch.arange(T, device=q.device).unsqueeze(1)
    ki = torch.arange(T, device=q.device).unsqueeze(0)
    causal    = ki <= qi                       # [T, T]
    in_window = causal & ((qi - ki) < window_size)

    mask = torch.where(in_window, 0.0,
           torch.where(causal, gate[:, :, None, :],
                       float("-inf")))

    return F.scaled_dot_product_attention(
        q, k, v, attn_mask=mask)
\end{lstlisting}

Listing~\ref{lst:spkv} provides a minimal PyTorch implementation of the
SP-KV attention mechanism described in Section~\ref{sec:gated-kv}.
The \texttt{UtilityPredictor} produces per-key, per-head utilities
$u_s^{l,k}\in(0,1)$. The \texttt{sp\_kv\_attention} function builds the
attention mask from three regions: the sliding window (bias${}=0$),
long-range positions (bias${}=\log u$ during training, $0$ or $-\infty$
at inference), and future positions (bias${}=-\infty$).
The listing uses multi-head attention ($H_q = H_{kv}$) for clarity.
Under grouped-query attention, the utility predictor produces one gate
per KV head; the attention kernel broadcasts each gate across the
corresponding query-head group.

\end{document}